\PassOptionsToPackage{dvipsnames,table}{xcolor}%
\documentclass[acmtog,screen,authorversion]{acmart} %

\acmSubmissionID{0203}

\usepackage{booktabs} %

\usepackage{array}
\newcolumntype{P}[1]{>{\centering\arraybackslash}p{#1}}
\newcolumntype{M}[1]{>{\centering\arraybackslash}m{#1}}
\def\doublehline{\hline\hline}
\def\thickhline{\noalign{\hrule height 1pt}}

\usepackage{pifont}
\newcommand{\cmark}{\ding{52}}%

\newcommand{\CHECKED}[1]{{{\color{black}{#1}}}} %
\newcommand{\NEWSA}[1]{{\color{black}{#1}}} %
\newcommand{\NEW}[1]{{\color{black}{#1}}} %
\newcommand{\NEWSACAM}[1]{{\color{black}{#1}}} %

\newcolumntype{o}{>{\columncolor{YellowGreen!60}}M}

\usepackage{import} %
\graphicspath{{figs/}}

\usepackage[ruled,noend]{algorithm2e} %

\SetAlFnt{\small}
\SetAlCapFnt{\small}
\SetAlCapNameFnt{\small}
\SetAlCapHSkip{0pt}
\IncMargin{-\parindent}
\usepackage{ifpdf}
\usepackage{graphicx}
\ifpdf
  
\else
\fi
\definecolor{cyan}{cmyk}{1.0,0,0,0.3}
\definecolor{darkgreen}{rgb}{0,0.5,0}
\definecolor{orange}{rgb}{1,0.5,0}
\definecolor{magenta}{cmyk}{0,1,0,0}
\definecolor{darkyellow}{cmyk}{0,0,0.75,0}
\definecolor{gray}{rgb}{0.8,0.8,0.8}

\newenvironment{tightitemize}{
\vspace{-1.5mm}
\begin{itemize}
  \setlength{\itemsep}{1pt}
  \setlength{\parskip}{2pt}
  \setlength{\parsep}{0pt}}{\end{itemize}

}
\usepackage{amsmath} %

\usepackage[toc,page]{appendix} %
\usepackage{wrapfig} %
\usepackage{comment}
\usepackage{bm}
\usepackage{cuted} %
\usepackage{lipsum} %
\usepackage{mathtools} %
\usepackage{algorithmicx}
\usepackage{algpseudocode}
\usepackage{nicefrac}

\usepackage{gensymb}
\usepackage{float}
\usepackage{soul}
\usepackage{array}
\usepackage{multirow}
\usepackage{hhline}
\usepackage{setspace}
\usepackage{nicefrac}

\algdef{SE}[DOWHILE]{Do}{doWhile}{\algorithmicdo}[1]{\algorithmicwhile\ #1}%

\makeatletter
\renewcommand{\ALG@beginalgorithmic}{\small}
\makeatother

\usepackage{datenumber}
\usepackage{calc}
\usepackage[mmddyyyy]{datetime}

\newcounter{datetoday}
\newcounter{diffyears}
\newcounter{diffmonths}
\newcounter{diffdays}

\newcommand{\difftoday}[3]{%
      \setmydatenumber{datetoday}{\the\year}{\the\month}{\the\day}%
      \setmydatenumber{diffdays}{#1}{#2}{#3}%
      \addtocounter{diffdays}{-\thedatetoday}%
      \ifnum\value{diffdays}>0
        \def\diffbefore{}%
        \def\diffafter{left}%
      \else
        \def\diffbefore{}%
        \def\diffafter{ago}%
        \setcounter{diffdays}{-\value{diffdays}}%
      \fi
      \setcounter{diffyears}{\value{diffdays}/365}%
      \setcounter{diffdays}{\value{diffdays}-365*\value{diffyears}}%
      \setcounter{diffmonths}{\value{diffdays}/30}%
      \setcounter{diffdays}{\value{diffdays}-30*\value{diffmonths}}%
      \diffbefore
      \ifnum\value{diffyears}=0
      \else
        \ifnum\value{diffyears}>1
            \thediffyears\space years,
        \else
            \thediffyears\space year,
        \fi
      \fi
      \ifnum\value{diffmonths}=0
      \else
        \ifnum\value{diffmonths}>1
            \thediffmonths\space months
        \else
            \thediffmonths\space month
        \fi
      \fi
      \ifnum\value{diffdays}=0
      \else
        \ifnum\value{diffdays}>1
            \thediffdays\space days
        \else
            \thediffdays\space day
        \fi
      \fi
      \diffafter
}

\usepackage[ruled]{algorithm2e} %

\SetAlFnt{\small}
\SetAlCapFnt{\small}
\SetAlCapNameFnt{\small}
\SetAlCapHSkip{0pt}
\graphicspath{{./figs/}}

\usepackage{microtype}
\acmJournal{TOG}

\setcopyright{rightsretained}
\acmJournal{TOG}
\acmYear{2024} 
\acmVolume{43} 
\acmNumber{6} 
\acmArticle{275} 
\acmMonth{12}
\acmDOI{10.1145/3687767}

\begin{document}
\title{\NEWSA{Polarimetric BSSRDF Acquisition of Dynamic Faces}}

\author{Hyunho Ha}
\orcid{0000-0002-8375-6449}
\affiliation{%
  \institution{KAIST}
  \country{South Korea}}
\email{hhha@vclab.kaist.ac.kr}

\author{Inseung Hwang}
\orcid{0000-0002-9971-1202}
\affiliation{%
  \institution{KAIST}
  \country{South Korea}}
\email{ishwang@vclab.kaist.ac.kr}

\author{Nestor Monzon}
\orcid{0000-0002-5323-3233}
\affiliation{%
 \institution{Universidad de Zaragoza - I3A}
 \country{Spain}}
\email{nmonzon@unizar.es}

\author{Jaemin Cho}
\orcid{0000-0003-2800-5105}
\affiliation{%
  \institution{KAIST}
  \country{South Korea}}
\email{jmcho@vclab.kaist.ac.kr}

\author{Donggun Kim}
\orcid{0000-0002-6670-6263}
\affiliation{%
  \institution{KAIST}
  \country{South Korea}}
\email{dgkim@vclab.kaist.ac.kr}

\author{Seung-Hwan Baek}
\orcid{0000-0002-2784-4241}
\affiliation{%
  \institution{POSTECH}
  \country{South Korea}}
\email{shwbaek@postech.ac.kr}  

\author{Adolfo Muñoz}
\orcid{0000-0002-8160-7159}
\affiliation{%
 \institution{Universidad de Zaragoza - I3A}
 \country{Spain}}
\email{adolfo@unizar.es}

\author{Diego Gutierrez}
\orcid{0000-0002-7503-7022}
\affiliation{%
  \institution{Universidad de Zaragoza - I3A}
  \country{Spain}}
\email{diegog@unizar.es}

\author{Min H. Kim}
\orcid{0000-0002-5078-4005}
\affiliation{%
  \institution{KAIST}
  \country{South Korea}}
\email{minhkim@kaist.ac.kr}

\begin{abstract}
\NEWSA{
Acquisition and modeling of polarized light reflection and scattering help reveal the shape, structure, and physical characteristics of an object, which is increasingly important in computer graphics.
However, current polarimetric acquisition systems are limited to static and opaque objects.
Human faces, on the other hand, present a particularly difficult challenge, given their complex structure and reflectance properties, the strong presence of spatially-varying subsurface scattering, and their dynamic nature.
We present a new polarimetric acquisition method for dynamic human faces, which focuses on capturing spatially varying appearance and precise geometry, across a wide spectrum of skin tones and facial expressions. It includes both single and heterogeneous subsurface scattering, index of refraction, and specular roughness and intensity, among other parameters, while revealing biophysically-based components such as inner- and outer-layer hemoglobin, eumelanin and pheomelanin. 
Our method leverages such components' unique multispectral absorption profiles to quantify their concentrations, which in turn inform our model about the complex interactions occurring within the skin layers.
To our knowledge, our work is the first to simultaneously acquire polarimetric and spectral reflectance information alongside biophysically-based skin parameters and geometry of dynamic human faces. 
Moreover, our polarimetric skin model integrates seamlessly into various rendering pipelines.
}

\end{abstract}



\begin{teaserfigure}
	\centering
	\vspace{5mm}
	\small%
	\graphicspath{{./figs/sa2024/teaser/}}%
	\sffamily%
	\resizebox{1.0\linewidth}{!}{
\begingroup%
  \makeatletter%
  \providecommand\color[2][]{%
    \errmessage{(Inkscape) Color is used for the text in Inkscape, but the package 'color.sty' is not loaded}%
    \renewcommand\color[2][]{}%
  }%
  \providecommand\transparent[1]{%
    \errmessage{(Inkscape) Transparency is used (non-zero) for the text in Inkscape, but the package 'transparent.sty' is not loaded}%
    \renewcommand\transparent[1]{}%
  }%
  \providecommand\rotatebox[2]{#2}%
  \newcommand*\fsize{\dimexpr\f@size pt\relax}%
  \newcommand*\lineheight[1]{\fontsize{\fsize}{#1\fsize}\selectfont}%
  \ifx\svgwidth\undefined%
    \setlength{\unitlength}{695.10489396bp}%
    \ifx\svgscale\undefined%
      \relax%
    \else%
      \setlength{\unitlength}{\unitlength * \real{\svgscale}}%
    \fi%
  \else%
    \setlength{\unitlength}{\svgwidth}%
  \fi%
  \global\let\svgwidth\undefined%
  \global\let\svgscale\undefined%
  \makeatother%
  \begin{picture}(1,0.29828395)%
    \lineheight{1}%
    \setlength\tabcolsep{0pt}%
    \put(0,0){\includegraphics[width=\unitlength,page=1]{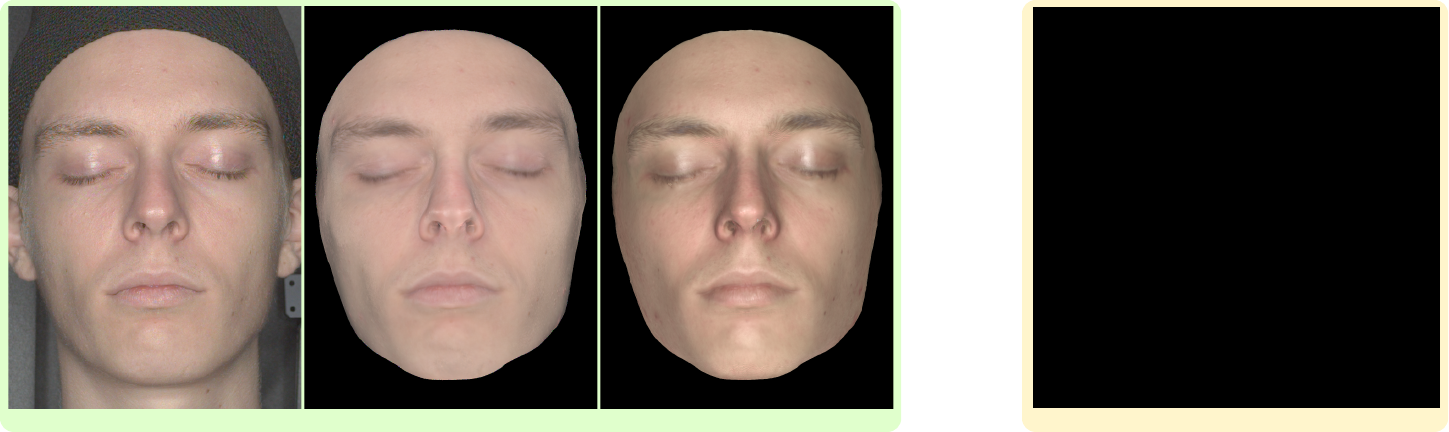}}%
    \put(0.85933616,0.00514092){\color[rgb]{0,0,0}\makebox(0,0)[t]{\lineheight{1.25}\smash{\begin{tabular}[t]{c}\small{Dynamic}\end{tabular}}}}%
    \put(0,0){\includegraphics[width=\unitlength,page=2]{teaser_P13_hwang_v2.pdf}}%
    \put(0.3116563,0.02192339){\color[rgb]{1,1,1}\makebox(0,0)[t]{\lineheight{1.25}\smash{\begin{tabular}[t]{c}\footnotesize{\citet{hwang_sparse_2022}}\end{tabular}}}}%
    \put(0.31729144,0.00514092){\color[rgb]{0,0,0}\makebox(0,0)[t]{\lineheight{1.25}\smash{\begin{tabular}[t]{c}\small{Static}\end{tabular}}}}%
    \put(0,0){\includegraphics[width=\unitlength,page=3]{teaser_P13_hwang_v2.pdf}}%
    \put(0.75192024,0.03661919){\color[rgb]{1,1,1}\makebox(0,0)[t]{\lineheight{1.25}\smash{\begin{tabular}[t]{c}\scriptsize{Melanin}\end{tabular}}}}%
    \put(0.82012619,0.03661919){\color[rgb]{1,1,1}\makebox(0,0)[t]{\lineheight{1.25}\smash{\begin{tabular}[t]{c}\scriptsize{Rel. eumel.}\end{tabular}}}}%
    \put(0.88834421,0.03661919){\color[rgb]{1,1,1}\makebox(0,0)[t]{\lineheight{1.25}\smash{\begin{tabular}[t]{c}\scriptsize{He. (outer)}\end{tabular}}}}%
    \put(0.95655028,0.03661919){\color[rgb]{1,1,1}\makebox(0,0)[t]{\lineheight{1.25}\smash{\begin{tabular}[t]{c}\scriptsize{He. (inner)}\end{tabular}}}}%
    \put(0.85358809,0.02192339){\color[rgb]{1,1,1}\makebox(0,0)[t]{\lineheight{1.25}\smash{\begin{tabular}[t]{c}\footnotesize{Biophysical parameters}\end{tabular}}}}%
    \put(0,0){\includegraphics[width=\unitlength,page=4]{teaser_P13_hwang_v2.pdf}}%
    \put(0.78119859,0.06400244){\color[rgb]{1,1,1}\makebox(0,0)[t]{\lineheight{1.25}\smash{\begin{tabular}[t]{c}\tiny{0.0}\end{tabular}}}}%
    \put(0.78119859,0.10473105){\color[rgb]{0,0,0}\makebox(0,0)[t]{\lineheight{1.25}\smash{\begin{tabular}[t]{c}\tiny{0.3}\end{tabular}}}}%
    \put(0.84940603,0.06394252){\color[rgb]{1,1,1}\makebox(0,0)[t]{\lineheight{1.25}\smash{\begin{tabular}[t]{c}\tiny{0.0}\end{tabular}}}}%
    \put(0.84940603,0.10473635){\color[rgb]{0,0,0}\makebox(0,0)[t]{\lineheight{1.25}\smash{\begin{tabular}[t]{c}\tiny{0.5}\end{tabular}}}}%
    \put(0,0){\includegraphics[width=\unitlength,page=5]{teaser_P13_hwang_v2.pdf}}%
    \put(0.98572788,0.14460826){\color[rgb]{0,0,0}\makebox(0,0)[t]{\lineheight{1.25}\smash{\begin{tabular}[t]{c}\tiny{0}\end{tabular}}}}%
    \put(0.98572788,0.19448009){\color[rgb]{0,0,0}\makebox(0,0)[t]{\lineheight{1.25}\smash{\begin{tabular}[t]{c}\tiny{$\pi$}\end{tabular}}}}%
    \put(0,0){\includegraphics[width=\unitlength,page=6]{teaser_P13_hwang_v2.pdf}}%
    \put(0.9176121,0.06394252){\color[rgb]{1,1,1}\makebox(0,0)[t]{\lineheight{1.25}\smash{\begin{tabular}[t]{c}\tiny{0.0}\end{tabular}}}}%
    \put(0.9176121,0.10473105){\color[rgb]{0,0,0}\makebox(0,0)[t]{\lineheight{1.25}\smash{\begin{tabular}[t]{c}\tiny{0.3}\end{tabular}}}}%
    \put(0.98581954,0.06400244){\color[rgb]{1,1,1}\makebox(0,0)[t]{\lineheight{1.25}\smash{\begin{tabular}[t]{c}\tiny{0.0}\end{tabular}}}}%
    \put(0.98581954,0.10479628){\color[rgb]{0,0,0}\makebox(0,0)[t]{\lineheight{1.25}\smash{\begin{tabular}[t]{c}\tiny{0.5}\end{tabular}}}}%
    \put(0.78280556,0.12833812){\color[rgb]{1,1,1}\makebox(0,0)[t]{\lineheight{1.25}\smash{\begin{tabular}[t]{c}\footnotesize{Heterogeneous}\end{tabular}}}}%
    \put(0.78280557,0.11919319){\color[rgb]{1,1,1}\makebox(0,0)[t]{\lineheight{1.25}\smash{\begin{tabular}[t]{c}\footnotesize{subsurface scattering}\end{tabular}}}}%
    \put(0.92534244,0.12402222){\color[rgb]{1,1,1}\makebox(0,0)[t]{\lineheight{1.25}\smash{\begin{tabular}[t]{c}\footnotesize{Angle of linear polarization}\end{tabular}}}}%
    \put(0.85280156,0.28384916){\color[rgb]{1,1,1}\makebox(0,0)[rt]{\lineheight{1.25}\smash{\begin{tabular}[t]{r}\footnotesize{Cross polarization}\end{tabular}}}}%
    \put(0.61323155,0.28384916){\color[rgb]{1,1,1}\makebox(0,0)[rt]{\lineheight{1.25}\smash{\begin{tabular}[t]{r}\footnotesize{Parallel polarization}\end{tabular}}}}%
    \put(0.40882419,0.28384916){\color[rgb]{1,1,1}\makebox(0,0)[rt]{\lineheight{1.25}\smash{\begin{tabular}[t]{r}\footnotesize{Parallel polarization}\end{tabular}}}}%
    \put(0.20441706,0.28384916){\color[rgb]{1,1,1}\makebox(0,0)[rt]{\lineheight{1.25}\smash{\begin{tabular}[t]{r}\footnotesize{Parallel polarization}\end{tabular}}}}%
    \put(0.10716286,0.02192339){\color[rgb]{1,1,1}\makebox(0,0)[t]{\lineheight{1.25}\smash{\begin{tabular}[t]{c}\footnotesize{Photograph}\end{tabular}}}}%
    \put(0.51804838,0.02192339){\color[rgb]{1,1,1}\makebox(0,0)[t]{\lineheight{1.25}\smash{\begin{tabular}[t]{c}\footnotesize{Ours}\end{tabular}}}}%
    \put(0,0){\includegraphics[width=\unitlength,page=7]{teaser_P13_hwang_v2.pdf}}%
    \put(0.84104753,0.17780703){\color[rgb]{1,1,1}\makebox(0,0)[t]{\lineheight{1.25}\smash{\begin{tabular}[t]{c}\tiny{Light strips}\end{tabular}}}}%
  \end{picture}%
\endgroup%
}%
	\vspace{-3mm}
	\caption[]{\label{fig:teaser}%
\NEWSA{Current polarimetric acquisition systems are limited to static, opaque objects. In this work, we present a novel appearance acquisition method that allows us to obtain biophysically-based polarimetric BSSRDF and surface geometry of dynamic faces. Our polarimetric appearance parameters include index of refraction, specular roughness, single scattering roughness, specular intensity, and single scattering intensity. We also introduce an end-to-end multispectral optimization with heterogeneous subsurface scattering, revealing biophysically-based skin parameters including inner- and outer-layer hemoglobin, eumelanin and pheomelanin. Our model is compatible with popular human skin models in graphics. Please refer to the supplemental video for dynamic results.}
	\vspace{5mm}
}	
\end{teaserfigure}

\maketitle

\section{Introduction}
\label{sec:intro}
\NEWSA{%
Polarization can provide valuable information about the shape and physical characteristics of an object. As a result, polarization-based systems have become increasingly important in computer graphics. These require capturing and processing four-dimensional Stokes vectors to account for all potential polarization states. 

Furthermore, the capture of polarimetric reflectance requires control over incident and outgoing light, which are both Stokes vectors, 
so the reflectance function is often represented by a four-by-four Mueller matrix that links each component of the incident light with its outgoing counterpart. The complexity of this matrix increases the computational cost of estimating polarization appearance parameters, requiring additional observations with different incoming and outgoing directions and polarization states. Ellipsometry is the most common technique to estimate this matrix, using structured optical measurements to characterize how light interactions affect the polarization state. Current methods to capture spatially varying polarimetric reflectance techniques need to apply strong assumptions to make the problem tractable. As a result, they are still limited to acquiring polarimetric appearance information from static opaque objects.

In this work, we lift these restrictions and capture the polarimetric reflectance of dynamic deformable objects with strong, spatially-varying subsurface scattering. We specifically target the challenging case of human faces, which are both deformable and translucent. Figure~\ref{fig:teaser} compares two different polarization rendering results by a state-of-the-art method~\cite{hwang_sparse_2022} and ours. By accounting for heterogeneous subsurface scattering, our model yields more precise results, closer to the photographic reference. 

Estimating such subsurface scattering is challenging since light scatters through the multiple translucent layers of skin. We base our subsurface scattering appearance model on   
biophysical components, in particular melanin (eumelanin and pheomelanin) and hemoglobin (oxy-hemoglobin and deoxy-hemoglobin). Our hardware setup captures six spectral observations, and we rely on the unique spectral absorption profiles of these biophysical components to estimate their individual contribution to the final appearance.

Our algorithm is made up of two stages. First, we capture polarimetric observations of the subject's static face from a wide range of angles, from which we optimize the appearance parameters that serve as initialization for the next stage. In the second stage, for each frame, we follow a similar optimization, starting from the outcome of the first stage. With this approach, we obtain per-frame high-quality dynamic geometry, as well as spatially-varying appearance parameters represented as texture maps. 
\NEWSACAM{We use polarization imaging for estimating specular reflectance, single scattering  and geometrical detail, while we approximate the biophysical parameters from the spectral observations of subsurface scattering.}
To our knowledge, our technique is the first to capture polarimetric reflectance on dynamic deformable objects.
We show results across a wide spectrum of skin tones and facial expressions (both in the main paper and supplemental material). Moreover, our polarimetric skin model integrates seamlessly into many existing rendering pipelines.
\NEWSACAM{Our code is available for research purposes\footnote{\url{https://github.com/KAIST-VCLAB/polarimetric-bssrdf-dynamic-face.git}}.}
}

\section{Related Work}
\label{sec:relatedwork}

\NEWSA{
\paragraph{Polarimetric imaging.} 

Polarimetric imaging has widely been used in computer graphics. Passive systems use cameras fitted with polarizers, positioned in front of the lens (e.g.,~\cite{miyazaki_polarization-based_2003, atkinson_recovery_2006, huynh_shape_2013,kadambi_polarized_2015, tozza_linear_2017, riviere_polarization_2017, zhu_depth_2019, cui_polarimetric_2019, deschaintre_deep_2021, cao_multi-view_2023}), or the image sensor~\cite{ba_deep_2020, lei_shape_2022,zhao_polarimetric_2022, dave_pandora_2022}, 
while active systems incorporate both polarized light sources and polarized cameras (e.g.,~\cite{ma_rapid_2007,ghosh_practical_2008,riviere_single-shot_2020,azinovic_high-res_2022,ghosh_circularly_2010}). 
In general, these setups only measure specific polarization states, such as linear polarization at particular angles or circular polarization.

Other works aim to capture polarimetric appearance across different polarization states \cite{baek_simultaneous_2018, baek_polarimetric_2021, baek_all-photon_2022}. \citet{hwang_sparse_2022} combined a polarization-array camera with a polarized flashlight.
However, these approaches are limited to static scenes and do not take subsurface scattering explicitly into account. In contrast, our work allows us to capture polarimetric information on dynamic faces, including the effects of subsurface scattering.
}

\paragraph{Face acquisition.} Numerous methods have been developed to acquire high-quality geometric shapes and the appearance of static faces \NEWSACAM{(e.g., ~\cite{debevec_acquiring_2000,weyrich_analysis_2006,ma_rapid_2007,legendre_efficient_2018,ghosh_circularly_2010,fyffe_single-shot_2010,fyffe_comprehensive_2011,fyffe_near-instant_2016,imai_multi-spectral_1998,shrestha_multispectral_2010,ghosh_practical_2008,azinovic_high-res_2022}).}
Since they all require multiple structured light patterns and/or input from various viewpoints, they are unsuitable for dynamic captures.

Dynamic face capture methods, on the other hand, often use passive illumination, taking images of objects under uniform lighting conditions. 
Multi-view camera systems rely on stereo matching for geometry acquisition~\cite{beeler_high-quality_2010}, or tracking in image space~\cite{bradley_high_2010, beeler_high-quality_2011}, but do not reconstruct the appearance of skin.
Monocular single-shot~\cite{tran_extreme_2018, sengupta_sfsnet_2018, tran_towards_2019, tran_learning_2019}, video sequences~\cite{garrido_reconstructing_2013, shi_automatic_2014, cao_real-time_2015, ichim_dynamic_2015}, or binocular video sequences~\cite{valgaerts_lightweight_2012} have been used to obtain both geometric information and appearance. These approaches typically assume simplified reflectance models of human skin, for instance, including only diffuse albedo or not taking subsurface scattering into account \cite{gotardo_practical_2018}.
\citet{riviere_single-shot_2020} developed a passive stereo-capture system to acquire specular reflectance and diffuse albedo. As opposed to our work, the method assumes a pre-determined subsurface scattering profile, while each frame needs to be processed independently for animated sequences.
Since single-shot input is ill-conditioned for human skin acquisition, recent research has turned to learning from active multi-view lighting systems~\cite{li_learning_2020,liu_rapid_2022,zhang_video-driven_2022,bi_deep_2021}. 
These learning-based methods are constrained by the training and test datasets, which do not describe the reflectance of human faces in a physically-based way. In contrast, our multispectral polarimetric subsurface scattering model yields approximate meaningful, spatially-varying, and time-resolved biophysically-based appearance parameters for dynamic faces.

\NEW{
\paragraph{Biophysical appearance acquisition}

Existing methods to approximate biophysical parameters of human skin usually rely on simplified models, such as assuming diffuse reflectance, not taking into account subsurface scattering, or not handling dynamic changes in appearance. 
\citet{tsumura1999independent,tsumura2003image} created an image-based method to separate the spatial patterns of melanin and hemoglobin in human skin through independent-component analysis of a skin color image. This model was later extended to take into account the more complex properties of skin~\cite{krishnaswamy2004biophysically} based on multispectral images~\cite{preece_spectral_2004, donner_layered_2008,chen2015hyperspectral}, RGB diffuse reflectance images~\cite{alotaibi_biophysical_2017,aliaga_estimation_2022}, \NEWSACAM{or different lighting conditions~\cite{gitlina_practical_2020,aliaga_hyperspectral_2023,li2024practical}}. Some of these methods require precomputed textures for inverse rendering, which may lead to visible discretization artifacts or rely on rendered datasets.
Regarding dynamic models, \citet{jimenez_practical_2010} presented a method focused on the acquisition of simplified hemoglobin maps from cross-polarizing filters, requiring multiple captures with the subject repeating the same movements. Later, \citet{Iglesias2015skin} introduced a statistically-based model of human skin that captured the time-varying effects of aging, as the structure of skin and its chromophores change over the years.
In contrast, our technique disambiguates the dynamic changes of biophysically-based components, such as oxy-hemoglobin, deoxy-hemoglobin, eumelanin or pheomelanin, as well as their full diffusion profiles, through a multispectral observation of subsurface scattering without requiring impractical repeated motions. Moreover, our system is capable of simulating appearance changes that occur within seconds, instead of decades. 
}

\section{Reflectance model of skin}
\label{sec:background}
We describe here the main aspects of our reflectance model for skin, including our polarimetric BSSRDF and biophysically-based parameters;
please refer to the supplemental document for additional details (\CHECKED{Supplemental Section~\ref{sec:polarization}}).

\subsection{Polarimetric BSSRDF Model}
\label{sec:pbssrdf_model}
A Stokes vector represents the polarization state of a light wave, and is denoted as $\mathbf{s}=[s_{0},s_{1},s_{2},s_{3}]^{\intercal}\in\mathbb{R}^{4\times1}$. Polarized  light $\mathbf{s}_{i}$ reflects off a surface as
	$\mathbf{s}_{o}=S\mathbf{P}(\boldsymbol{\omega}_{i}, \boldsymbol{\omega}_{o})\mathbf{s}_{i}$,
where $S={(\mathbf{n}\cdot{\boldsymbol{\omega}_{i}})}/{\Gamma^{2}}$ is the shading term with attenuation,
$\Gamma$ is the distance between the light source and the surface, and $\mathbf{P}(\boldsymbol{\omega}_{i}, \boldsymbol{\omega}_{o})$ is the polarimetric reflectance model that yields a Mueller matrix for incoming $\boldsymbol{\omega}_{i}$ and outgoing $\boldsymbol{\omega}_{o}$ directions \cite{wilkie_polarised_2012}.

Subsurface scattering describes how light enters a surface at point~$\mathbf{x}_{i}$ and exits at a different point~$\mathbf{x}_{o}$. Different from existing polarimetric reflectance models~\cite{baek_simultaneous_2018,baek_polarimetric_2021,hwang_sparse_2022},  we explicitly take into account 
\textit{heterogeneous} subsurface scattering. 
This is important for human skin since, although light becomes depolarized during multiple scattering, it gets polarized again when transmitted back out, thus becoming an additional source of reflectance information. 

\NEW{Our polarimetric reflectance model can then be expressed as
$\mathbf{P}=\mathbf{P}_{s}+\mathbf{P}_{ss}+\mathbf{P}_{sss}$,
where $\mathbf{P}_{s}$, $\mathbf{P}_{ss}$, and $\mathbf{P}_{sss}$ represent the specular, single scattering, and subsurface scattering components, respectively. 
We adopt the specular and single scattering terms from the recent state-of-the-art model by~\citet{hwang_sparse_2022} (\CHECKED{Supplemental Section~\ref{sec:pbrdf_model}}), and expand the model with our new subsurface scattering term $\mathbf{P}_{sss}$, described in the following paragraphs. 
}

\paragraph{Subsurface scattering}
We model 
 subsurface scattering $\mathbf{P}_{sss}$ from a diffusion profile $\rho_{sss}(||\mathbf{x}_{i}-\mathbf{x}_{o}||) $ as

\small
\begin{equation}
    \label{eq:p_sss}
	\begin{aligned}
	\mathbf{P}_{sss} ~ 
	&\resizebox{.85\hsize}{!}{$
	=\sum\limits_{\mathbf{x}_{i}\in\mathcal{S}}\mathbf{C}_{n\rightarrow o}(-\tilde{\phi}_{o})\mathbf{F}^{\mathcal{T}}(\mathbf{x}_{o},\theta_{o};\eta_{o})\mathbf{D}(\rho_{sss}(||\mathbf{x}_{i}-\mathbf{x}_{o}||))$} \\ 
	&\hspace{10mm}\resizebox{.37\hsize}{!}{$
	\cdot \mathbf{F}^{\mathcal{T}}(\mathbf{x}_{i},\theta_{i};\eta_{i})\mathbf{C}_{i\rightarrow n}(\tilde{\phi}_{i})$} \\
	&=\sum^{}_{\mathbf{x}_{i}\in\mathcal{S}}\rho_{sss}(||\mathbf{x}_{i}-\mathbf{x}_{o}||) 
	\resizebox{.6\hsize}{!}{$
	\left[
	\begin{array}{cccc}
		\mathcal{T}_{i}^{+}\mathcal{T}_{o}^{+} & -\mathcal{T}_{i}^{-}\mathcal{T}_{o}^{+}\xi_{i} & -\mathcal{T}_{i}^{-}\mathcal{T}_{o}^{+}\zeta_{i} & 0 \\
		-\mathcal{T}_{i}^{+}\mathcal{T}_{o}^{-}\xi_{o} & \mathcal{T}_{i}^{-}\mathcal{T}_{o}^{-}\xi_{i}\xi_{o} & \mathcal{T}_{i}^{-}\mathcal{T}_{o}^{-}\zeta_{i}\xi_{o} & 0 \\
		-\mathcal{T}_{i}^{+}\mathcal{T}_{o}^{-}\zeta_{o} & \mathcal{T}_{i}^{-}\mathcal{T}_{o}^{-}\xi_{i}\zeta_{o} & \mathcal{T}_{i}^{-}\mathcal{T}_{o}^{-}\zeta_{i}\zeta_{o} & 0 \\
		0 & 0 & 0 & 0
	\end{array}
	\right],
	$}
	\end{aligned}
\end{equation}
\normalsize

\noindent {where $\mathbf{C}_{n\rightarrow o}(-\tilde{\phi}_{o})$ and $\mathbf{C}_{i\rightarrow n}(\tilde{\phi}_{i})$ are the coordinate conversion matrices, $\mathbf{D}$ is the depolarization matrix, }$\mathbf{F}^{\mathcal{T}}$ is the Mueller matrix form of the Fresnel transmission coefficients $\mathcal{T}$
that takes into account the different effects on light polarized along the plane of incidence (${\parallel}$) and perpendicular to it (${\perp}$).
The $+/-$ superscript operators refer to $\mathcal{T}^{+}=(\mathcal{T}^{\perp}+\mathcal{T}^{\parallel})/2$ and $\mathcal{T}^{-}=(\mathcal{T}^{\perp}-\mathcal{T}^{\parallel})/2$.
Here $\zeta$ and $\xi$ are $\sin(2\phi)$ and $\cos(2\phi)$ of 
the polarimetric azimuth angle $\phi$ between the light frame and the interaction plane, and $\tilde{\phi}_{\{i,o\}}=\phi_{\{i,o\}}-\pi/2$ is the corresponding rotation angle.
The sum $\sum^{}_{\mathbf{x}_{i}\in\mathcal{S}}$ takes into account the fact that
all incoming points $\mathbf{x}_i$ of the surface $\mathcal{S}$ contribute to the outgoing illumination. 

The diffusion profile $\rho_{sss}$ is obtained from the absorption and scattering coefficients of two layers, \NEWSA{which depend on a set of biophysically-based parameters as described in Section~\ref{subsec:facelayer}}, by applying the multipole approximation~\cite{donner_light_2005}. We assume that the parameters vary slowly 
relative to the mean free path of light, and therefore are locally homogeneous.
For efficiency purposes, per-layer profiles are approximated to weighted sums of separable Gaussian functions~\cite{deon_efficient_2007, donner_layered_2008}. In the rest of the paper, we omit the dependance of $\rho_{sss}$ on the distance $||\mathbf{x}_{i}-\mathbf{x}_{o}||$ for the sake of brevity.

\subsection{Biophysically-Based Model}
\label{subsec:facelayer}

\begin{figure}[t]
	\centering
	\def\svgscale{1}
	\graphicspath{{./figs/absorption/}}
	\sffamily
	\resizebox{0.95\linewidth}{!}{
		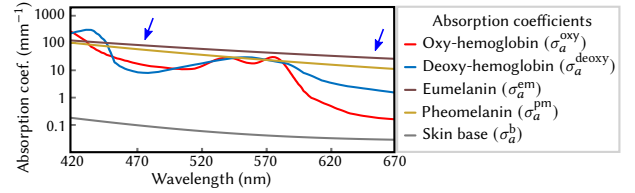}%
	\vspace{-3mm}%
	\caption[]{\label{fig:absorption}%
		Spectral absorption coefficients of each biophysical component in the human skin: oxy-hemoglobin, deoxy-hemoglobin, eumelanin, pheomelanin, and skin base.}
\end{figure}

Similar to other works~\cite{donner_spectral_2006, donner_layered_2008, jimenez_practical_2010}, we adopt a two-layer model, 
where each layer is characterized by its absorption and reduced scattering coefficients.  Figure~\ref{fig:absorption} shows the spectral absorption coefficients of each component, included in our model as explained in the next paragraphs. 

\paragraph{Outer-layer absorption}
The spectral absorption coefficient of the outer layer $\sigma_{a}^{\textrm{out}}$ is mainly due to the presence of melanin in the epidermis, and, to a lesser extent, to the presence of hemoglobin in the upper dermis, and is defined as~\cite{donner_layered_2008, jimenez_practical_2010}
\begin{align}
	\label{eq:epi_ab}
	\sigma_{a}^{\textrm{out}}=&C_{\text{m}}(\beta_{\text{m}}\sigma_{a}^{\text{em}}+(1-\beta_{\text{m}})\sigma_{a}^{\text{pm}})\\
	&+C_{\text{h,out}}(\gamma_\text{out}\sigma_{a}^{\text{oxy}}+(1-\gamma_\text{out})\sigma_{a}^{\text{deoxy}})
	+(1-C_{\text{m}}-C_{\text{h,out}})\sigma_{a}^{\text{b}},\nonumber
\end{align}
where $C_{\text{m}}$ and $C_{\text{h,out}}$ are the fractions of melanin and hemoglobin in the outer layer, respectively,
$\beta_{\text{m}}$~is the fraction of eumelanin in melanin, $\gamma_\text{out}$~is the oxy-hemoglobin fraction in hemoglobin. 
The values of 
{$C_{\text{m}}$, $C_{\text{h,out}}$, and $\beta_{\text{m}}$} 
are {estimated} from our captured data
while $\gamma_\text{out}$ is a constant
{(see Section~\ref{sec:optimization-biophysical}).} 
The different spectral absorption coefficients~$\sigma_{a}$ of eumelanin (em), pheomelanin (pm), oxy-hemoglobin (oxy), deoxy-hemoglobin (deoxy), and base (b) are given by previous work~\cite{prahl_optical_1999,jacques_skin_1998}.

\begin{figure*}[tp]
	\centering
	\def\svgscale{1}%
	\sffamily%
	\graphicspath{{./figs/system/}}%
	\resizebox{1.0\linewidth}{!}{
		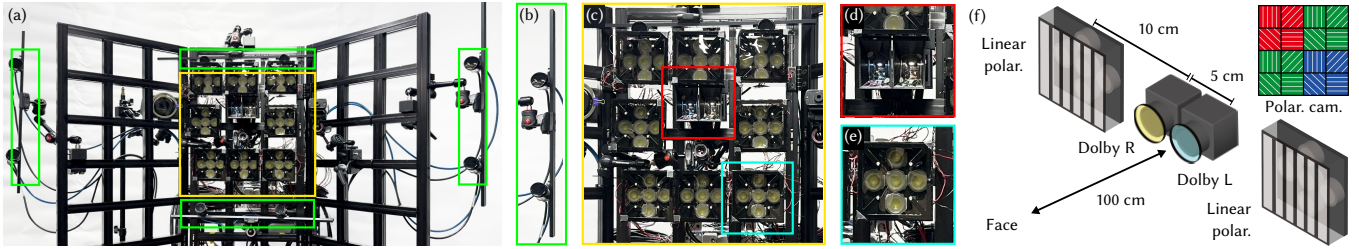}%
	\vspace{-2mm}%
	\caption[]{\label{fig:hardware}%
	\NEW{Our multispectral polarimetric imaging setup. 
	(a) Our system consists of a multispectral polarization module (yellow box) that captures four different linear polarization angles with six multispectral channels and four 3D stereo imaging modules (green boxes).
	(b) A closeup image of the stereo module composed of two machine vision cameras.  
	Stereo pairs are used to acquire dense depth maps to obtain a complete geometry model and its corresponding texture mapping for each frame.
	(c)--(e) Closeup images of the multispectral polarization camera module 
	equipped with two polarization cameras (d) covered with two different Dolby filters and
	40 LED lights, separated into eight modules of five LEDs, each shown in (e).
	Our multispectral polarization module captures six different multispectral channels using two different Dolby filters for three RGB channels.
	Each light is covered with a vertical polarization filter.
	(f)~Schematic diagram of the system configuration. 
	 The module also captures four different orientations of linear polarization. 
	 The eight light modules are placed at a distance of approximately 10\,cm. 
	 As the subject is placed 100\,cm away, light and camera become almost coaxial at a 5.72$^\circ$ angle.
	These cameras are synchronized through GPIO cables.
	We capture images at 20\,fps.}
	}
\end{figure*}

\paragraph{Inner-layer absorption}
The inner layer consists of a rich network of capillaries containing hemoglobin in the dermis. Its absorption coefficient $\sigma_{a}^{\textrm{in}}$ is mainly explained by hemoglobin, and can be expressed as~\cite{donner_spectral_2006}
\begin{align}
	\label{eq:dermis_ab}
	\sigma_{a}^{\textrm{in}}=&C_{\text{h,in}}(\gamma_\text{in}\sigma_{a}^{\text{oxy}}+(1-\gamma_\text{in})\sigma_{a}^{\text{deoxy}})
	+(1-C_{\text{h,in}})\sigma_{a}^{\text{b}},
\end{align}
where $C_{\text{h,in}}$ is the fraction of the hemoglobin in the inner layer, and $\gamma_\text{in}$ is its oxy-hemoglobin fraction. 
{The value of $C_{\text{h,in}}$ is also estimated from our captured data.} 
We assume that the fractions of oxy-hemoglobin to hemoglobin in the inner and outer layers are the same and have a fixed value $\gamma = \gamma_\text{in} = \gamma_\text{out} = 0.75$ as other existing models \cite{donner_layered_2008, jimenez_practical_2010}.

\paragraph{Reduced scattering}
The spectral reduced scattering coefficient of the outer layer $\sigma _s^{\textrm{out}'}$ 
{at wavelength $\lambda$~nm} is defined as~\cite{bashkatov_optical_2005}
\begin{equation}
	\label{eq:reduced_scatter}
	\sigma _s^{\textrm{out}'}(\lambda )_{} = 14.74 \times \lambda _{}^{ - 0.22} + 2.2 E^{11} \times \lambda _{}^{ - 4}.
\end{equation}
\NEWSA{The reduced scattering coefficient of the inner layer is 50$\%$ of the outer scattering coefficient, so it does not need to be explicitly estimated.}

\section{Multispectral Polarimetric Imaging}

\label{sec:system-design}
We summarize here the main aspects of our hardware design and polarimetric image formation model, and refer the reader to the supplemental material (\CHECKED{Supplemental Section~\ref{sec:polar-imaging}}) for more details. 

\subsection{Capture Hardware}
\label{subsec:hardware}

\begin{figure}[tp]
	\centering
	\def\svgscale{1}%
	\graphicspath{{./figs/calib/}}%
	\sffamily%
	\footnotesize%
	\resizebox{0.95\linewidth}{!}{
		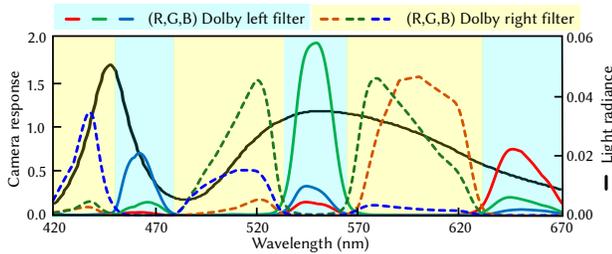}%
	\vspace{-4mm}%
	\caption[]{\label{fig:responsefunction}%
	Multispectral-channel calibration \NEWSACAM{and light spectral power distribution (black line)}. We disassemble left and right glass lenses from Dolby multispectral 3D anaglyph glasses to use each as a bandpass filter in front of two sRGB cameras. Each color channel is subdivided into two sub-color channels, resulting in six multispectral channels.
	}
\end{figure}

Our capture system is shown in Figure~\ref{fig:hardware}. 
It consists of a multispectral, polarimetric module in the center, and four additional 3D imaging modules surrounding it. 
The polarimetric imaging module is composed of two polarization machine vision cameras (BFS-U3-51SPC-C), synchronized at 20\,fps, each of them fitted with a different multispectral filter from off-the-shelf Dolby 3D glasses. Each polarization camera captures four linearly polarized components (0$\degree$, 45$\degree$, 90$\degree$, 135$\degree$) \NEWSACAM{at 2448$\times$2048 resolution. Note that conventional polarimetric cameras have a lower SNR and resolution than color cameras.} \NEW{The cameras are surrounded by {40} linearly-polarized $\sim$1500 lumen LED light sources (CXA-1512 \NEWSACAM{6500K, operating at 350mA with 36V}) {(eight modules of five LEDs each)} in a near-coaxial setup.}
Each Dolby 3D glass further filters the wavelength range of each of the camera's conventional red, green, and blue filters, effectively halving the range for each channel (the left camera captures the higher half, while the right one captures the lower one), yielding a coverage of the whole spectrum for a total of six samples 
{(see Figure~\ref{fig:responsefunction})}. This is particularly useful for the spectral response of human skin, in which the spectral absorption profiles of its components are identifiably different (Figure~\ref{fig:absorption}). 
Finally, each 3D imaging module consists of two machine vision cameras to capture dynamic 3D geometry. All cameras are synchronized
(\CHECKED{Supplemental Section~\ref{sec:calib-detail}}).

\subsection{Polarimetric Image Formation}
\label{sec:imageformation}
\NEW{Existing polarimetric acquisition methods can capture diffuse and specular information \cite{baek_image-based_2020,baek_simultaneous_2018} or even single scattering \cite{hwang_sparse_2022}, assuming that target objects are opaque.
We introduce a novel
subsurface scattering term (Equation~\eqref{eq:p_sss}) to handle translucency, as in human skin.

Since our light sources and cameras are in a near-coaxial setup, we can apply sparse ellipsometry algebraic simplifications to our polarimetric BSSRDF model~\cite{hwang_sparse_2022}. Given our linearly polarized captured images from each camera $I_{0},I_{90},I_{45},I_{135}$, we compute the following observations: 
\begin{tightitemize} 
\item The unpolarized subsurface scattering observation is defined as $I_{sss}=2I_{90}=S\sum_{\mathbf{x}_{i}\in\mathcal{S}}\rho_{sss}\mathcal{T}^{++}$.
\item The polarized subsurface scattering observation is defined as $I_{\zeta}=I_{135}-I_{45}=S\sum_{\mathbf{x}_{i}\in\mathcal{S}}\rho_{sss}\mathcal{T}^{-+}\zeta$.
\item The specular-dominant polarization observation is defined as $I_{s}=I_{0}-I_{90}=S(\bar{\kappa}_{s,ss}\mathcal{R}^{+}-\sum_{\mathbf{x}_{i}\in\mathcal{S}}\rho_{sss}\mathcal{T}^{-+}\xi)$. 
Here, $\bar{\kappa}_{s,ss}$ is the summation of the specular reflection term ${\kappa}_{s}$ and the single scattering reflection term ${\kappa}_{ss}$.
It contains a combination of specular reflection, single scattering, and multiple subsurface scattering.
\end{tightitemize}
 
 \NEW{Here,}
$\mathcal{R}^{+}=(\mathcal{R}^{\perp}+\mathcal{R}^{\parallel})/2$ represent Fresnel reflection coefficients,
$\mathcal{T}^{++}=\mathcal{T}^{+}\mathcal{T}^{+}$ is the multiplication of the positive Fresnel transmission coefficients, 
and $\mathcal{T}^{-+}=\mathcal{T}^{-}\mathcal{T}^{+}$ is the multiplication of the negative/positive coefficients.
\NEWSA{Please refer to the supplemental document for the complete mathematical details of our polarimetric image formation model (\CHECKED{Supplemental Section~\ref{sec:polar-image-formation-detail}}).}

We acquire these observations per frame and use them as input to our optimization algorithm (described in the next section), to obtain our full dynamic data.}

\section{Reconstruction of Dynamic Skin Appearance and Face Geometry}
\label{sec:optimization}

\begin{figure}
\centering
\def\svgscale{0.78}
\graphicspath{{./figs/}}
\resizebox{1\linewidth}{!}{
	\begin{normalsize}
		\sffamily%
	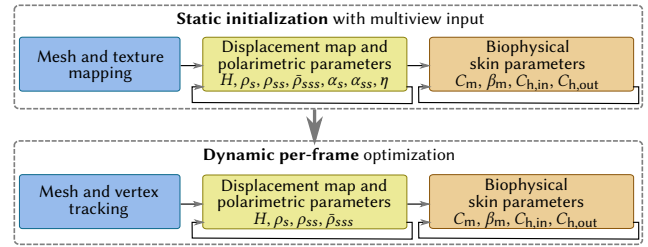
\end{normalsize}}%
\vspace{-3mm}%
\caption{\label{fig:overview}
Overview of our reconstruction algorithm. It consists of two stages: first, the initialization takes multiple views of a static face (top). We then perform a per-frame optimization from a single view (bottom). Both stages follow a similar procedure: first, we obtain a mesh from stereo pairs (left), then we iteratively optimize a displacement map and a set of polarimetric appearance parameters (middle), and lastly, we use inverse rendering to iteratively optimize the biophysical parameters of human skin (right).}
\vspace{-4mm}
\end{figure}

Our reconstruction algorithm consists of two stages (see Figure~\ref{fig:overview} for an overview). 
	The first stage is a \textit{static initialization} from multiple views of the same face, rotated thanks to a revolving chair. We first obtain the face's mesh from stereo pairs (Section~\ref{sec:geometry-tracking}); we then simultaneously optimize the displacement map plus the polarimetric appearance parameters (Section~\ref{sec:optimization-polarimetric}), then optimize the biophysical parameters (Section~\ref{sec:optimization-biophysical}). At each iteration, the face is rendered in order to calculate the loss function with respect to the input observations. The second stage is a \textit{dynamic per-frame optimization}, in which, starting with the results of the first stage, we optimize all the necessary parameters for every frame of the captured video in a similar manner as the first stage.

In more detail, given our hardware setup, from the four linearly polarized images ($I_{0},I_{90},I_{45},I_{135}$) and the four stereo view pairs, we obtain six-channel multispectral observations $I_{sss}$, $I_{\zeta}$ and $I_{s}$ through our image formation model (Section \ref{sec:imageformation}). We next search the geometric correspondences of spatially varying dynamic appearance changes over time. We aim to couple polarimetric appearance, biophysically-based skin parameters and the surface geometry to track appearance parameters on a face over time. In particular, the index of refraction $\eta$ (which affects all Fresnel coefficients), albedo ($\rho_{s}$ and $\rho_{ss}$) and roughness ($\alpha_{s}$ and $\alpha_{ss}$) of the specular (\emph{s}) and single scattering (\emph{ss}) components, and the diffusion profile $\rho_{sss}$, expressed as a function of biophysical parameters ($C_\textrm{m}$, $\beta_\textrm{m}$, $C_\textrm{h,in}$, $C_\textrm{h,out}$). 

All these parameters, plus a displacement map $H$ encoding geometry details, are spatially varying and are represented as texture maps over the skin's surface.
In the following, we explain each stage in more detail and refer the reader to the supplemental material for additional details (\CHECKED{Supplemental Section~\ref{sec:optimization-details}}).

\subsection{Geometry Reconstruction and Tracking}
\label{sec:geometry-tracking}

At both stages (static initialization and dynamic per-frame optimization), we first estimate the base mesh geometry using stereo matching~\cite{beeler_high-quality_2010, lipson_raft-stereo_2021} and Poisson surface reconstruction~\cite{kazhdan_poisson_2006}. With this initial geometry, we apply cylindrical texture mapping so that all the spatially-varying appearance parameters and small geometry variations (encoded as displacement map $H$) are modeled as textures. 
Note that previous work~\cite{hwang_sparse_2022} optimized vertices and their normals directly, requiring an additional Poisson reconstruction process at every iteration, which resulted in a blurrier geometry. By directly optimizing geometric details in the form of a displacement map $H$, we obtain
detailed, more accurate geometrical reconstructions. 
$H$ is optimized together with the polarimetric appearance parameters (Section~\ref{sec:optimization-polarimetric}). We assign $uv$ texture coordinates during the initialization stage after the first mesh has been optimized; in the subsequent per-frame optimization we track vertices, but texture coordinates remain unchanged. As a result, textures remain stable along frames, which improves the convergence of our optimization.

During the dynamic per-frame optimization stage, we additionally track corresponding vertices from the initial mesh via optical flow.
We stabilize the tracking by progressively computing the weighted average of the per-frame tracked motion with an anchor-based approach~\cite{beeler_high-quality_2011}, which yields the final per-frame mesh with stable texture coordinates per vertex.

\subsection{Optimization of Polarimetric Appearance Parameters}
\label{sec:optimization-polarimetric}
After optimizing the mesh, we optimize the spatially-varying appearance parameters that are related to polarization: index of refraction~$\eta$, the albedo $\rho_{s}$ and roughness $\alpha_{s}$ of the specular component, the albedo~$\rho_{ss}$ and roughness $\alpha_{ss}$ of the single scattering component, and a multiple scattering albedo $\bar{\rho}_{sss}$, which is a rough approximation of the diffusion profile $\rho_{sss} (||\mathbf{x}_{i}-\mathbf{x}_{o}||)$. This value will be later refined to a full diffusion profile when estimating the face's biologically-based parameters (Section~\ref{sec:optimization-biophysical}). We also optimize the displacement map~$H$, which provides the high-frequency details of the geometry. Parameters $\eta$, $\alpha_{s}$ and $\alpha_{ss}$ remain constant in time, and thus they only need to be optimized in the initialization stage. 

\NEWSACAM{Inspired by \citet{gotardo_practical_2018}, this initialization stage consists on rotating the static face, assuming parameter consistency across frames. As shown in previous work~\cite{nagano_skin_2015}, roughness might temporally vary when the skin is stretched or becomes sweaty, but estimating both roughness parameters $\alpha_{s}$ and $\alpha_{ss}$ from a single view at each frame is an ill-posed problem. The small errors coming from this assumption are compensated by albedos $\rho_{s}$ and $\rho_{ss}$ and the geometrical variations coming from the displacement map $H$, all of which are estimated per frame.}

In particular, we minimize the following energy function:
\begin{equation}\label{eq:p-inverse-main-loss}
\min_{\eta, \alpha_{s}, \alpha_{ss}, \rho_{s}, \rho_{ss}, \bar{\rho}_{sss}, H}\lambda_{\psi}\mathcal{L}_{\psi}+\lambda_{sss}\mathcal{L}_{sss}+\lambda_{s}\mathcal{L}_{s}+\lambda_{\phi}\mathcal{L}_{\phi}+\mathcal{L}_{\text{reg}},
\end{equation}
where $\mathcal{L}_{sss}$ is our 
subsurface scattering loss, 
$\mathcal{L}_{\psi}$ is the refractive index loss, $\mathcal{L}_{s}$ is the specular and single scattering loss, $\mathcal{L}_{\phi}$ is the azimuthal loss,  $\mathcal{L}_{\text{reg}}$ is the regularization term, and $\lambda_{\psi}=0.002$, $\lambda_{sss}=1$, $\lambda_{s}=1$, $\lambda_{\phi}=1$ are the corresponding loss weights. 
\NEW{We inherit the specular, single scattering, refractive index loss functions from \citet{hwang_sparse_2022} and the regularization term (that accounts for spatial and temporal coherency) from \citet{riviere_single-shot_2020}.
Previous work~\cite{hwang_sparse_2022} solves these loss terms by alternating the optimization of the refractive index loss ($\mathcal{L}_{\psi}$) and the specular and single scattering loss ($\mathcal{L}_{s}$) with azimuthal loss ($\mathcal{L}_{\phi}$), using a sequential quadratic programming algorithm. In contrast, we use a backward gradient descent-based method that minimizes losses simultaneously. Moreover, while current techniques are limited to polarimetric appearance of static and opaque objects, we capture translucency effects in our initialization stage and track dynamic changes during optimization \NEWSA{(\CHECKED{Supplemental Section~\ref{sec:optimization-details}}).} }

\paragraph{Subsurface scattering loss}
We formulate $\mathcal{L}_{sss}$ by comparing the rendered subsurface scattering image $\hat{I}^{t}_{sss}$ at time $t$ with the captured image $I^{t}_{sss}$ as
$\mathcal{L}_{sss}=\sum\nolimits_{t}\text{V}^{t}(\hat{I}^{t}_{sss}-I^{t}_{sss})^{2}$, where $\text{V}^{t}$ is the visibility texture map at frame $t$ for each view.
Optimizing the full diffusion profile $\rho_{sss}$ along with the rest of the variables is both computationally expensive and ill-conditioned. Therefore, as anticipated earlier, we account for a single multiple scattering albedo $\bar{\rho}_{sss}$ to approximate all the observations of subsurface scattering effects.
Since Fresnel transmittance of human skin does not change rapidly along the surface, we approximate the subsurface scattering reflectance as $I^{t}_{sss}=S\bar{\rho}_{sss}\mathcal{T}^{++}$. Once $\bar{\rho}_{sss}$ and the rest of the polarimetric appearance parameters are optimized, we obtain the full diffusion profile by optimizing the rest of the biophysical parameters, as explained in Section~\ref{sec:optimization-biophysical}.

\subsection{Optimization of Biophysically-based Parameters}
\label{sec:optimization-biophysical}
To estimate a full diffusion profile ${\rho}_{sss}$ from a multispectral observation of subsurface scattering $\bar{\rho}_{sss}$, we rely on the spectral profiles of the absorption coefficients of oxy-hemoglobin, deoxy-hemoglobin, eumelanin, pheomelanin, and the skin base parameter.
Since their absorption coefficients are different with respect to their spectral structure (see Figure~\ref{fig:absorption}), we can leverage our multispectral measurements (Figure~\ref{fig:responsefunction}) to disambiguate the concentrations of the different biophysical components of our skin model ($C_{\text{m}}$, $\beta_{\text{m}}$, $C_{\text{h,in}}$, and $C_{\text{h,out}}$). 

We minimize the photometric loss between $\bar{\rho}_{sss}$ and the rendered subsurface scattering using the full diffusion profile. The main challenge is the differentiation of the diffusion profile 
with respect to the biophysical parameters, since forward optimization is neither efficient nor scalable for high-resolution textures  \cite{donner_layered_2008}.
To tackle the lack of end-to-end derivatives, we propose a coordinate descent method~\cite{wright_coordinate_2015} using alternating least squares. Our optimization is thus split into two subproblems: first, obtaining the weights of the Gaussians that define $\bar{\rho}_{sss}$; second, estimating the biophysical parameters of $\bar{\rho}_{sss}$.

We discretize the spectral absorptions into fifteen multispectral channels. To calculate the photometric loss, we convert these channels into our camera's six channels using our system's spectral calibration functions. 
For efficiency, early-stage iterations are calculated at a coarser resolution.
We use nine Gaussians to approximate each profile and merge this subsurface scattering with the contributions from the specular and single scattering components to render the full appearance model. 
This efficient method enables fast gradient descent iterations to approximate the skin's biophysical parameters (\CHECKED{Supplemental Section~\ref{sec:sss-optimization}}).

\begin{figure*}[!t]
	\centering
	\vspace{-5mm}%
	\def\svgscale{1}%
	\sffamily%
	\footnotesize%
	\resizebox{0.87\linewidth}{!}{
		\graphicspath{{./figs/sa2024/results/}}
		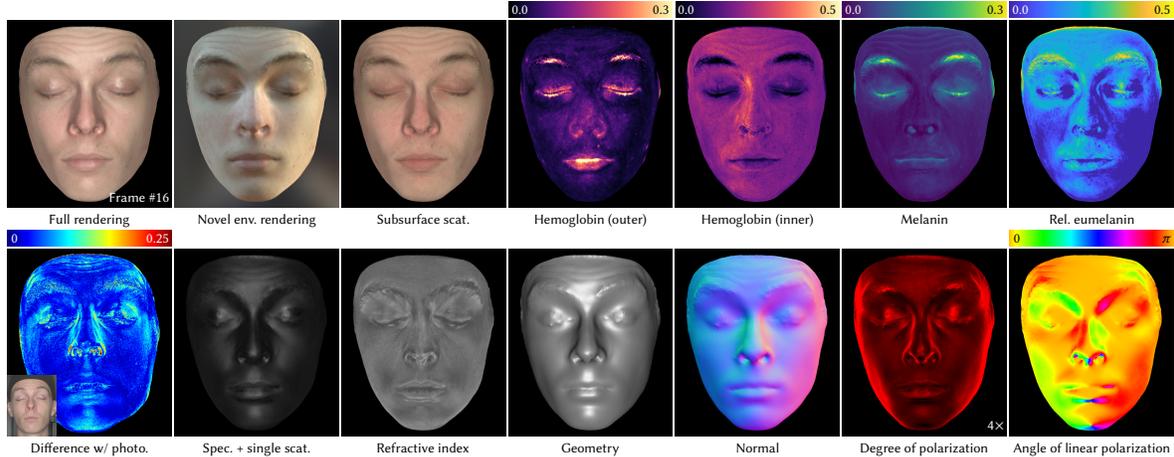}%
	\vspace{-4mm}%
	\caption[]{\label{fig:dynamic-results}%
		\NEWSA{Our dynamic face reconstruction results. We record data from 11 individuals, each exhibiting diverse skin tones, gender identities, and ethnic backgrounds. Our method successfully captures polarimetric reflectance parameters, biophysical parameters, and geometry with high accuracy. Refer to the supplemental document (Appendix A) 
		 and video for more results of dynamic faces.}}
\end{figure*}

\newcolumntype{o}{>{\columncolor{YellowGreen!60}}M}
\section{Results and Validation}
\label{sec:results}
\NEWSA{We estimate a per-texel polarimetric BSSRDF that consists of a three-by-three Mueller matrix of linear polarization.
Note that our reflectance function is nine times larger than the conventional BSSRDF.}
Our code runs on a machine equipped with an AMD EPYC 7763 CPU of 2.45\,GHz and an NVIDIA A100 GPU. \NEW{In the first stage of our method, polarimetric appearance optimization takes around 180 minutes with 200 frames,
 while biophysical multispectral optimization takes 50 minutes.
In the second stage, polarimetric optimization takes around 180 minutes with 50 frames,
 and biophysical optimization takes about 20 minutes.

We illustrate the versatility of our reconstructions on eleven subjects with different skin tones, genders, and ethnicities, performing various dynamic facial expressions.}
\NEWSACAM{Our method requires a near-coaxial light-camera configuration, with the subject's face positioned at the optical center of our polarimetric imaging unit during the capture.}
In our first initialization stage, participants maintain a neutral facial expression while the lighting and viewpoint angles change by spinning their heads. In the second stage, we instruct the participants to perform various facial expressions.
\NEWSA{Note that we use a \textbf{completely novel viewpoint of an RGB} camera to render and validate the results.}
Please refer to the \NEWSA{supplemental video} for \NEWSA{the capturing environments of our setup}. 
Figure~\ref{fig:dynamic-results} and 
\NEWSA{the appendix of the supplemental document}
show, for each subject, the resulting polarimetric BSSRDF, biophysical skin parameters, refractive index, normals, and geometry. These parameters present distinct variations across different subjects, according to their skin tone.

\begin{figure}[!t]
	\centering
	\def\svgscale{1}
	\sffamily%
	\scriptsize%
	\resizebox{0.92\linewidth}{!}{
		\graphicspath{{./figs/sa2024/bp_validation/}}
		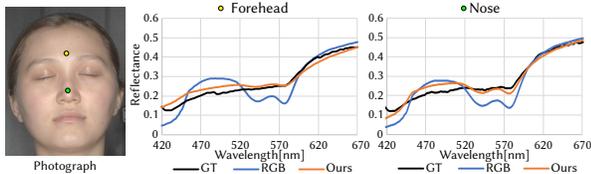}%
	\vspace{-4mm}%
	\caption[]{\label{fig:bp_validation}%
		\NEW{Validation of our multispectral reflectance. Our multispectral system enables accurate reconstructions of spectral reflectance compared to the ground-truth measurements.
		Results using trichromatic channels exhibit larger deviations from the ground truth.}}
	\vspace{-2mm}%
\end{figure}

\NEWSA{\paragraph{Validation}
We first validate our results comparing our estimated spectral reflectance with the ground-truth reflectance measured by a hyperspectral camera (SpecIM).
Figure~\ref{fig:bp_validation} shows how our reconstruction results closely match the ground-truth measurements.
In addition, to validate the accuracy of the refractive index that our method estimates, 
we compare our estimated refractive indices with the reference refractive indices of spherical objects, measured by their Brewster angles~\cite{baek_image-based_2020}. 
As shown in Table~\ref{tb:valid_ref_idx}, our system can measure the refractive indices of objects with high accuracy.}

\begin{figure}[!t]
	\centering
	\def\svgscale{1}%
	\sffamily%
	\small%
	\resizebox{0.89\linewidth}{!}{
		\graphicspath{{./figs/sa2024/mueller/}}
		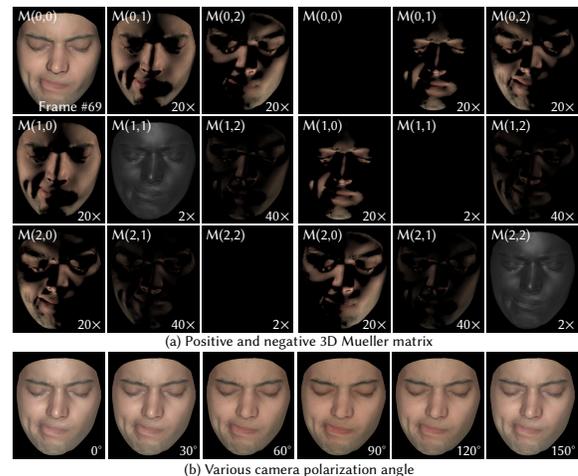}%
	\vspace{-4mm}%
	\caption[]{\label{fig:mueller}%
		\NEWSA{Our polarimetric reflectance results. (a) Positive and negative values of polarimetric reflectance function as a 3D Mueller matrix. We scale the intensity of each image for visualization purposes. (b) Polarization rendering simulates various polarimetric reflections of linear polarization angles for the camera.}}
	\vspace{-1mm}%
\end{figure}

\newcommand{\darkgreen}{\color{darkgreen}{\cmark}}
\begin{table}[!t]
	\small
	\centering
	\caption{\label{tb:valid_ref_idx} 
Validation on refractive index measurements. We compare our estimations $\eta_\textrm{ours}$ with known refractive indices $\eta_\textrm{gt}$ of ten real-world objects. We achieve a high accuracy with the mean reconstruction error of $0.028$.}
\vspace{-3mm}
\resizebox{0.78\linewidth}{!}{%
	\begin{tabular}{cc|l| c c c}
		\cline{2-6}
		\multirow{11}{*}{\parbox{0.17\linewidth}{\resizebox{1.0\linewidth}{!}{
					\sffamily%
					\graphicspath{{./figs/s2024/ior_validation/}}%
					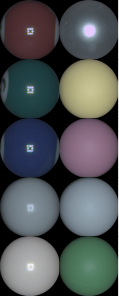}}}%
		&Object & Material & $\eta_{\textrm{gt}}$ & $\eta_{\textrm{ours}}$ & Diff. \\
		\cline{2-6}
		&1& Red billiard 			&1.485 	&1.446 	&0.038 \\
		&2& Green billiard 			&1.469	&1.516	&0.047 \\
		&3& Blue billiard 			&1.504	&1.503	&0.001 \\
		&4& White billiard 			&1.463	&1.410	&0.053 \\
		&5& POM 					&1.462	&1.447	&0.015 \\ 
		&6& Fake pearl 				&2.295	&2.263	&0.032 \\ 
		&7& Yellow silicone 		&1.303	&1.297	&0.005 \\ 
		&8& Pink silicone 			&1.177	&1.211	&0.034 \\ 
		&9& White silicone 			&1.248	&1.272	&0.024 \\ 
		&10& Light green silicone 	&1.343	&1.311	&0.032 \\ 
		\cline{2-6}
	\end{tabular}
}%
\end{table}

\NEW{\paragraph{Polarimetric reflectance}
\NEWSA{Our work is the first to capture polarimetric reflectance functions of human faces in the form of the $3\times3$ Mueller matrices, as shown in Figure~\ref{fig:mueller}(a)}.
\NEWSA{It allows us to simulate polarimetric face appearance changes by the linear polarization angle changes on the camera (Figure~\ref{fig:mueller}(b)).}
Also, this enables us to explore other polarization metrics, such as the angle of linear polarization (AoLP) or the degree of polarization (DoP), as shown in the second row of Figure~\ref{fig:dynamic-results}. We also show the captured index of refraction, which is crucial for our optimization and is only obtainable thanks to this polarimetric information.}

\begin{figure}[!t]
	\centering
	\vspace{0mm}%
	\def\svgscale{1}
	\sffamily%
	\resizebox{0.9\linewidth}{!}{
		\graphicspath{{./figs/s2024/dop_validation/}}
		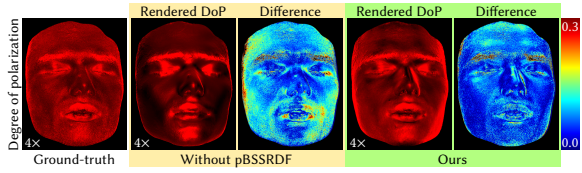}%
	\vspace{-3mm}%
	\caption[]{\label{fig:dop_validation}%
		\NEW{Validation of our polarimetric reflectance. We compare the degree of polarization of our estimated pBSSRDF with that of the ground-truth measurement by a polarization camera.}}
	\vspace{-2mm}%
\end{figure}

\begin{table}[!t]
	\footnotesize
	\centering
	\caption{\label{tb:ablation_spec} 
		\NEWSA{Ablation study of the impact of each component in the specular-dominant polarization observation. We average the RMSE values over 200 different novel camera poses with 11 different participants.}}
	\vspace{-4mm}
	\begin{tabular}
		{M{0.25\linewidth} M{0.2\linewidth} M{0.2\linewidth} M{0.18\linewidth}}
		\hline
		Refractive index and azimuthal loss & Single scattering parameters & Refractive index parameter & Average RMSE \\
		\hline
		\darkgreen & - & - &  $4.44{\cdot}10^{-2}$\\
		\darkgreen & \darkgreen & - &  $4.13{\cdot}10^{-2}$ \\
		\darkgreen & \darkgreen & \darkgreen &  $\mathbf{3.94{\cdot}10^{-2}}$ \\
		\hline
	\end{tabular}
	\vspace{-2mm}
\end{table}

\NEWSA{Existing methods of face capture~\cite{gotardo_practical_2018,riviere_single-shot_2020,azinovic_high-res_2022} use a single refractive index value being assumed for the entire face region.}
\NEWSA{In contrast, we optimize spatially-varying refractive-index values as well as polarimetric appearance parameters, which are critical to achieve accurate reconstruction of polarimetric appearance as shown in Figure~\ref{fig:dop_validation}.}
\NEWSA{The skin on the face has varying concentrations of oil and moisture, with higher levels on the forehead and nose compared to the cheeks or lips. 
In order to validate our pBSSRDF measurement, we compare the degree of polarization of our polarimetric rendering with that of the ground-truth measurement captured by a reference polarization camera. 
Our polarimetric rendering with the estimated pBSSRDF demonstrates a strong agreement with the ground-truth measurement of the degree of polarization, showing high accuracy.}
\NEWSA{Moreover, as shown in Table~\ref{tb:ablation_spec}, leveraging the polarimetric loss term with spatially varying single scattering and refractive index parameters results in the minimum RMSE error on the specular-dominant polarization observation.}

\begin{figure}[!t]
	\centering
	\vspace{0mm}%
	\def\svgscale{1}%
	\footnotesize%
	\sffamily%
	\graphicspath{{./figs/s2024/multispec/}}
	\resizebox{1.0\linewidth}{!}{
		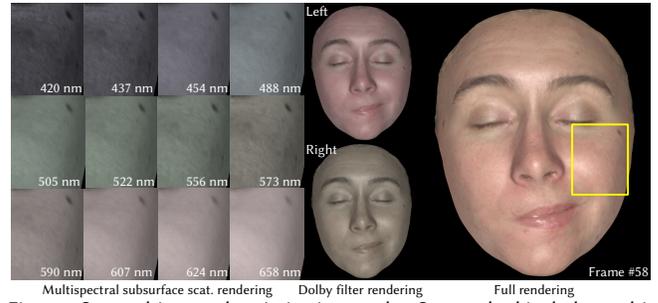}%
	\vspace{-4mm}%
	\caption[]{\label{fig:multispec}%
		\NEW{Our multispectral optimization results. Our method includes multispectral subsurface-scattering rendering across 15 different wavelengths, revealing wavelength-dependent scattering characteristics, like enhanced texture detail at shorter wavelengths. 
		We generate RGB images from these multispectral images using Dolby left/right-filter responses, supplementing the standard sRGB image. }}
\end{figure}

\NEWSA{\paragraph{Multispectral optimization}
Figure~\ref{fig:multispec} shows our multispectral optimization results rendered from our estimated biophysically-based parameters.
It can be seen how, as expected, sharper details can be recovered at shorter wavelengths, since longer wavelengths scatter further inside the skin~\cite{donner_spectral_2006}.} 

\begin{figure}[!t]
	\centering%
	\vspace{0mm}%
	\small%
	\sffamily%
	\def\svgscale{1}%
	\resizebox{0.99\linewidth}{!}{
		\graphicspath{{./figs/s2024/albedo_changes/}}
		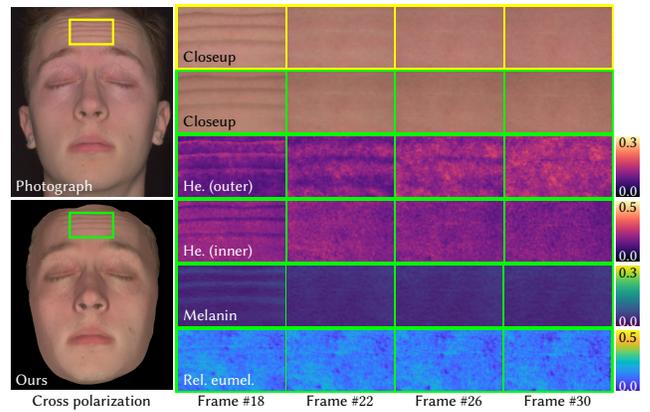}%
	\vspace{-3mm}%
	\caption[]{\label{fig:albedo_changes}%
		\NEW{Dynamic parameter changes. Our captured biophysical parameter map shows both blood flow changes and melanin map changes, caused by the wrinkles, being consistent with the dynamic appearance of the forehead.}}
	\vspace{0mm}
\end{figure}

\begin{figure}[!t]
	\centering
	\vspace{0mm}%
	\def\svgscale{1}
	\sffamily%
	\small%
	\resizebox{1.0\linewidth}{!}{
		\graphicspath{{./figs/bloodchange/}}
		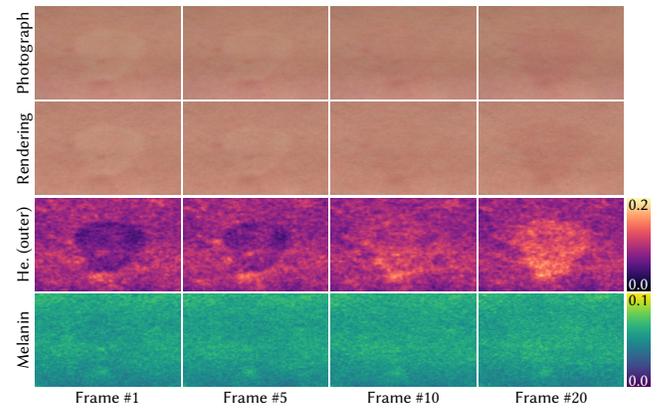}%
	\vspace{-3mm}%
	\caption[]{\label{fig:blood-flow}%
		Blood flow alterations in the forehead region after removing applied pressure. Hemoglobin concentrations in the outer layer return to their previous level as blood re-enters the pressed region.}
	\vspace{0mm}
\end{figure}

\NEWSA{Moreover, our method can handle dynamic changes in the appearance of different nature. Figure~\ref{fig:albedo_changes} shows time-varying changes in the distribution of the biophysically-based parameters caused by wrinkles in the forehead, while Figure~\ref{fig:blood-flow} illustrates changes in the hemoglobin concentration due to applied pressure. At Frame~\#1, such concentration is lower around the pressed region, but after approximately two seconds, blood re-enters the area (see Frames~\#10 and \#20). As expected, there is no significant change in melanin concentration.}

\begin{figure*}[!ht]
	\centering
	\vspace{-4mm}%
	\def\svgscale{1}%
	\footnotesize%
	\sffamily%
	\graphicspath{{./figs/sa2024/comparison/P4/}}
	\resizebox{1.0\linewidth}{!}{
		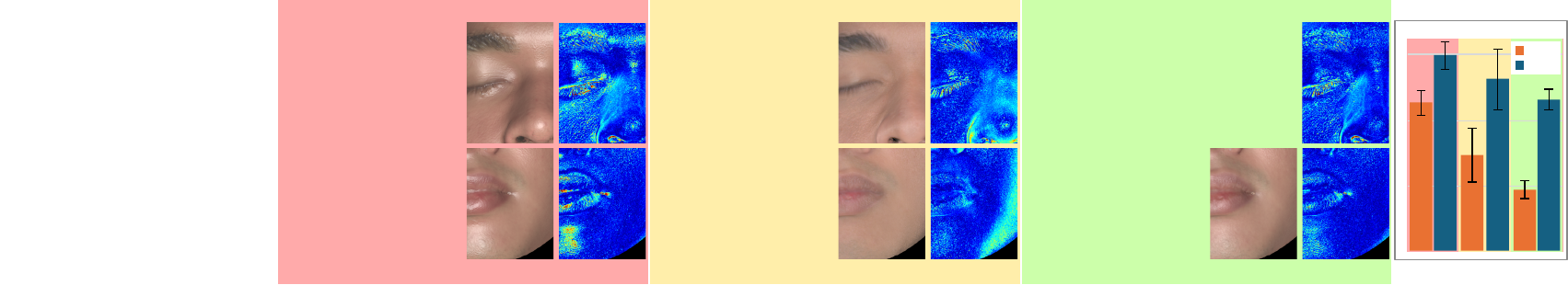}%
	\vspace{-4mm}%
	\caption[]{\label{fig:comparison}%
		\NEWSA{Comparison of appearance with state-of-the-art face acquisition model~\cite{riviere_single-shot_2020} \NEWSACAM{using our implementation} and static polarimetric acquisition~\cite{hwang_sparse_2022}. \citet{riviere_single-shot_2020} model over-estimated the specular highlights due to the absence of spatially-varying specular roughness, single scattering, and refractive index. \citet{hwang_sparse_2022} fails to reconstruct the specular highlights in the forehead and nose (yellow) and baked the specularity in the cheek (green) due to their alternative and cluster-based optimization scheme.
			Our method successfully reconstructs the highlight details of specular reflection in the nose and forehead region, together with detailed pore-level appearance. We also compute the average RMSE of the 200 different poses on a novel RGB camera on 11 participants, as shown in the right plot. Our method achieves the smallest reconstruction error on both specular and full rendering images.}}
\end{figure*}

\NEW{\paragraph{Comparison with prior works}
We directly compare our results with the recent, state-of-the-art face acquisition method of~\citet{riviere_single-shot_2020}, and the polarimetry method of~\citet{hwang_sparse_2022} with static scenes as shown in Figure~\ref{fig:comparison}.
\NEWSACAM{Comparisons with the work by \citet{riviere_single-shot_2020} are difficult, since there is no publicly available code or dataset. Therefore, we have implemented their method based on our own framework, increasing the number of input images from twelve in the original work to 200, to provide a more fair comparison.}

\NEWSA{As shown in the figure, assuming homogeneous specular roughness and refractive index makes \citet{riviere_single-shot_2020} overestimate specular highlights. On the other hand, \citet{hwang_sparse_2022} underestimates them due to their alternative, cluster-based optimization scheme. Our reconstruction of specular highlights and overall reflectance is more accurate, thanks to spatially varying specular roughness, single scattering, refractive index, and joint optimization. We also compute the RMSE of the specularity and the full rendering (right bar plot) images in 200 different poses of the 11 participants. Our method gives the smallest RMSE value on both images.}}

\begin{figure}[!t]
	\centering
	\def\svgscale{1}
	\sffamily%
	\small%
	\resizebox{1.0\linewidth}{!}{
		\graphicspath{{./figs/sa2024/geo/}}
		\begingroup%
  \makeatletter%
  \providecommand\color[2][]{%
    \errmessage{(Inkscape) Color is used for the text in Inkscape, but the package 'color.sty' is not loaded}%
    \renewcommand\color[2][]{}%
  }%
  \providecommand\transparent[1]{%
    \errmessage{(Inkscape) Transparency is used (non-zero) for the text in Inkscape, but the package 'transparent.sty' is not loaded}%
    \renewcommand\transparent[1]{}%
  }%
  \providecommand\rotatebox[2]{#2}%
  \newcommand*\fsize{\dimexpr\f@size pt\relax}%
  \newcommand*\lineheight[1]{\fontsize{\fsize}{#1\fsize}\selectfont}%
  \ifx\svgwidth\undefined%
    \setlength{\unitlength}{303.35650346bp}%
    \ifx\svgscale\undefined%
      \relax%
    \else%
      \setlength{\unitlength}{\unitlength * \real{\svgscale}}%
    \fi%
  \else%
    \setlength{\unitlength}{\svgwidth}%
  \fi%
  \global\let\svgwidth\undefined%
  \global\let\svgscale\undefined%
  \makeatother%
  \begin{picture}(1,0.35374317)%
    \lineheight{1}%
    \setlength\tabcolsep{0pt}%
    \put(0.16501843,0.00510645){\makebox(0,0)[t]{\lineheight{1.25}\smash{\begin{tabular}[t]{c}(a) \NEWSACAM{Generic stereo matching}\end{tabular}}}}%
    \put(0.49284723,0.00510645){\makebox(0,0)[t]{\lineheight{1.25}\smash{\begin{tabular}[t]{c}(b) \citet{hwang_sparse_2022}\end{tabular}}}}%
    \put(0.83498153,0.00510645){\makebox(0,0)[t]{\lineheight{1.25}\smash{\begin{tabular}[t]{c}(c) Ours\end{tabular}}}}%
    \put(0,0){\includegraphics[width=\unitlength,page=1]{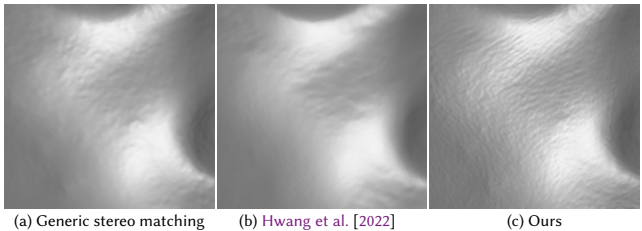}}%
  \end{picture}%
\endgroup%
}%
	\vspace{-2mm}%
	\caption[]{\label{fig:geo}%
		\NEWSA{Comparison of geometry reconstruction. By leveraging multiview polarimetric information in our optimization of the displacement map, our method yields more precise geometric reconstructions. (a) \NEWSACAM{Generic stereo matching (similar to \citet{beeler_high-quality_2010}'s method without mesoscopic augmentation)}. (b) Poisson-based inverse rendering~\cite{hwang_sparse_2022}. (c) Our method.}}
\end{figure}

\NEWSA{In terms of geometric accuracy, 
Figure~\ref{fig:geo} shows how our approach leads to artifact-free, more detailed reconstructions than previous approaches using \NEWSACAM{generic} stereo-matching~\cite{beeler_high-quality_2010} \NEWSACAM{without mesoscopic augmentation}, or Poisson-based inverse rendering optimization~\cite{hwang_sparse_2022}.}

\begin{figure}[!t]
	\centering
	\vspace{0mm}%
	\def\svgscale{1}%
	\footnotesize%
	\sffamily%
	\graphicspath{{./figs/sa2024/structure_light/}}
	\resizebox{\linewidth}{!}{
		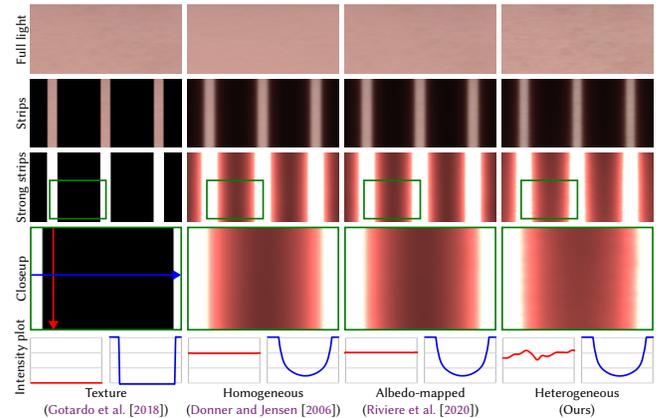}%
	\vspace{-3mm}%
	\caption[]{\label{fig:sss_strip}%
		\NEWSA{Full and strip light rendering results of texture-based~\cite{gotardo_practical_2018}, homogeneous subsurface scattering-based~\cite{donner_spectral_2006}, homogeneous subsurface scattering with albedo-map based on~\cite{riviere_single-shot_2020}, and our heterogeneous subsurface scattering method using estimated parameter maps. Each plot shows the intensity variation across the vertical cross-section line (red) and the horizontal line (blue).}}
\end{figure}

\NEWSA{Moreover, Figure~\ref{fig:sss_strip} compares the impact of our two-layer heterogeneous model as similar to that of \citet{donner_layered_2008}, by using structured light patterns. Previous methods based on diffuse albedo~\cite{gotardo_practical_2018, hwang_sparse_2022} cannot accurately simulate subsurface scattering in human skin. Although homogeneous subsurface scattering models~\cite{donner_spectral_2006} combined with albedo-mapped models~\cite{riviere_single-shot_2020} can simulate subsurface scattering, they cannot clearly depict the heterogeneity of the spatially varying parameters, as demonstrated in our method.}

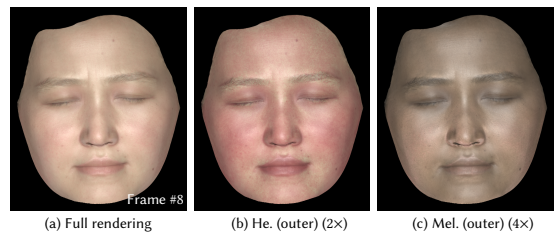
\begin{figure}[!t]
	\centering
	\vspace{0mm}%
	\def\svgscale{1}
	\resizebox{0.85\linewidth}{!}{
		\graphicspath{{./figs/s2024/editing/}}%
		\sffamily%
		\begingroup%
  \makeatletter%
  \providecommand\color[2][]{%
    \errmessage{(Inkscape) Color is used for the text in Inkscape, but the package 'color.sty' is not loaded}%
    \renewcommand\color[2][]{}%
  }%
  \providecommand\transparent[1]{%
    \errmessage{(Inkscape) Transparency is used (non-zero) for the text in Inkscape, but the package 'transparent.sty' is not loaded}%
    \renewcommand\transparent[1]{}%
  }%
  \providecommand\rotatebox[2]{#2}%
  \newcommand*\fsize{\dimexpr\f@size pt\relax}%
  \newcommand*\lineheight[1]{\fontsize{\fsize}{#1\fsize}\selectfont}%
  \ifx\svgwidth\undefined%
    \setlength{\unitlength}{322.4352787bp}%
    \ifx\svgscale\undefined%
      \relax%
    \else%
      \setlength{\unitlength}{\unitlength * \real{\svgscale}}%
    \fi%
  \else%
    \setlength{\unitlength}{\svgwidth}%
  \fi%
  \global\let\svgwidth\undefined%
  \global\let\svgscale\undefined%
  \makeatother%
  \begin{picture}(1,0.41403812)%
    \lineheight{1}%
    \setlength\tabcolsep{0pt}%
    \put(0.1616181,0.00629845){\makebox(0,0)[t]{\lineheight{1.25}\smash{\begin{tabular}[t]{c}(a) Full rendering\end{tabular}}}}%
    \put(0.5138621,0.00630063){\makebox(0,0)[t]{\lineheight{1.25}\smash{\begin{tabular}[t]{c}(b) He. (outer) ($2\times$)\end{tabular}}}}%
    \put(0.85227391,0.00630063){\makebox(0,0)[t]{\lineheight{1.25}\smash{\begin{tabular}[t]{c}(c) Mel. (outer) ($4\times$)\end{tabular}}}}%
    \put(0,0){\includegraphics[width=\unitlength,page=1]{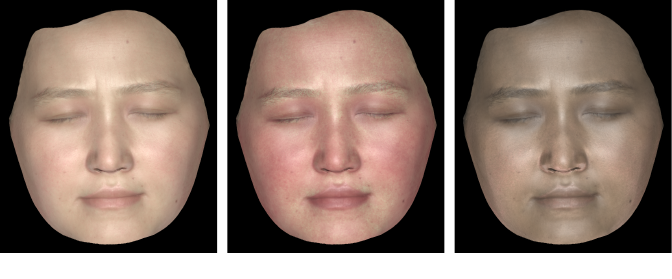}}%
    \put(0.31919384,0.05030959){\color[rgb]{1,1,1}\makebox(0,0)[rt]{\lineheight{1.25}\smash{\begin{tabular}[t]{r}Frame $\#8$\end{tabular}}}}%
  \end{picture}%
\endgroup%
}%
	\vspace{-3mm}%
	\caption[]{\label{fig:editing}%
		\NEW{Editing face biophysical parameters. (a)~Rendering with the original parameters. 
		(b)~Increased outer layer hemoglobin ($2\times$). 
		(c)~Increased melanin~($4\times$).}}
	\vspace{-2mm}%
\end{figure}

\NEW{\paragraph{Face appearance editing}
Lastly, our model enables the editing and the exploration of the effect of the different components on the skin's final appearance. Figure~\ref{fig:editing} shows how changes in hemoglobin and melanin affect such appearance. As expected, when the hemoglobin concentration in the outer layer increases, the subject's skin tone becomes reddish, while an increase in melanin leads to a more tanned appearance.}

\section{Discussion and Limitations}
\label{sec:discussion}
\NEWSA{We have presented a novel polarimetric imaging system that obtains 3D geometry and polarimetric reflectance of dynamic deformable surfaces made of translucent materials, for the particular case of human faces. The system is comprised of multispectral polarization cameras, polarized light sources and stereo imaging modules. Our skin BSSRDF model is two-layered, based on the main biophysically-based  components, which we approximate through a multispectral optimization based on their distinctive spectral profiles.}

\NEWSACAM{Our system's spatial resolution is half (2K) of conventional machine vision cameras (4K). We anticipate that the spatial resolution of BSSRDFs can be significantly improved when higher-resolution polarimetric cameras become available in the future.
Additionally, our off-the-shelf Dolby filters have spectral overlaps at specific wavelengths (570nm--620nm, Figure~\ref{fig:responsefunction}); the three-channel sRGB colors converted from our six-channel multispectral optimization via linear color transformation might thus exhibit a subtle error. Theoretically, this could be fixed by using custom bandpass filters that do not overlap.
Last, we do not explicitly include ambient occlusion to make the optimization manageable. As a consequence, for high-frequency geometry variations ambient occlusion shading may be misinterpreted as reflectance, which in turn may lead to small, local melanin variations.}
	
\NEWSA{Our optimization obtains a set of parameter maps, both polarimetric and biophysically-based. While we have validated the accuracy of some of the obtained parameters (index of refraction), we cannot claim that each individual parameter (particularly, the biophysically-based ones) is fully accurate. Still, the resulting global polarimetric appearance is a good match w.r.t. the input, and the behavior of each component is plausible. }

\begin{figure}[hbp]
	\centering
	\vspace{0mm}%
	\def\svgscale{1}
	\sffamily%
	\footnotesize%
	\resizebox{1.0\linewidth}{!}{
		\graphicspath{{./figs/sa2024/limitation/}}
		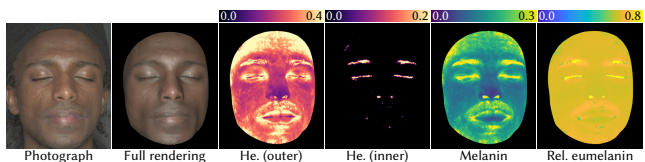}%
	\vspace{-3mm}%
	\caption[]{\label{fig:limitation}%
\NEWSA{Limitation example with very dark skin. While our algorithm successfully reconstructs both the geometry and appearance, it underestimates the contribution of the inner layer of hemoglobin
because most of the incident light gets absorbed by melanin and eumelanin present in the outer layer.}}
	\vspace{-4mm}
\end{figure}

\NEW{We have shown results across a wide spectrum of skin tones. However, we have noticed that inner layer components, especially hemoglobin, may be underestimated for subjects with very dark skin (Figure~\ref{fig:limitation}).
\NEWSA{This is because estimating non-invasive in-vivo biophysical parameters relies on the energy returned from the skin. When such energy is low, both existing methods and commercial products~\cite{shi_accuracy_2022,fawzy_racial_2022} may fail. This is therefore a common issue affecting very dark skin tones.}
Our methodology, nevertheless, could be applied to different multi-layered appearance models with different compositions, which is an interesting avenue for future work.
Moreover, applying other spectral illumination setups~\cite{preece_spectral_2004,gitlina_practical_2020,aliaga_hyperspectral_2023} is an exciting exploration for future research.}

\begin{acks}
Min H.~Kim acknowledges the Korea NRF grant (RS-2024-00357548), the MSIT/IITP of Korea (RS-2022-00155620, RS-2024-00398830, 2022-0-00058, and 2017-0-00072), Microsoft Research Asia, LIG, and Samsung Electronics.
Nestor Monzon acknowledges a Gobierno de Aragon predoctoral grant (2023–2027).
\end{acks}

\bibliographystyle{ACM-Reference-Format}
\bibliography{face-bibliography}
\clearpage


\onecolumn
\appendix
\section*{\textbf{Supplemental Document:\\
Polarimetric BSSRDF Acquisition of Dynamic Faces}}
\vspace{2cm}

\title[Supplemental Document: Polarimetric BSSRDF Acquisition of Dynamic Faces]{Supplemental Document: \\ Polarimetric BSSRDF Acquisition of Dynamic Faces}

\newcolumntype{o}{>{\columncolor{YellowGreen!60}}M}

\section{Related Work Overview}
\label{sec:related-work}
We summarize the contributions of related methods for face acquisition in Table~\ref{tb:relatedworks}.

\section{Polarization and subsurface scattering}
\label{sec:polarization}

\subsection{Stokes-Mueller Formalism}
\label{sec:stokes-mueller}
A Stokes vector represents the polarization state of a light wave and is denoted as $\mathbf{s}=[s_{0},s_{1},s_{2},s_{3}]^{\intercal}\in\mathbb{R}^{4\times1}$. 
The elements of the Stokes vector include: $s_{0}=L$, the intensity of the light; $s_{1}=L\psi\cos2\varsigma\cos2\chi$, and $s_{2}=L\psi\sin2\varsigma\cos2\chi$, the power of the $0\degree$ and $45\degree$ linear polarization components, respectively; and $s_{3}=L\psi\sin2\chi$, the power of the right circular polarization component. 
$\varsigma$ is the polarization angle, $\chi$ is the ellipticity angle, and $\psi=\sqrt{s_{1}^{2}+s_{2}^{2}+s_{3}^{2}}/s_{0}$ is the degree of polarization (DoP), defined as the ratio of the magnitude of the polarized vector elements to the intensity of the light. The effect on the polarization of the interaction between light and any element can be represented by a Mueller matrix $\mathbf{M}\in\mathbb{R}^{4\times4}$, that transforms a Stokes vector $\mathbf{s}_{\text{in}}$ into $\mathbf{s}_{\text{out}}$ as $\mathbf{s}_{\text{out}}=\mathbf{M}\mathbf{s}_{\text{in}}$.
For a complete description of polarized light, see the works of~\citet{collett2005field} and~\citet{wilkie_polarised_2012}.

\subsection{Fresnel Equation}
\label{sec:fresnel-eq}
The change of Stokes vectors by the transmission and reflection of light can be represented using the Fresnel Mueller matrix $\mathbf{F}^{F\in\{\mathcal{T},\mathcal{R}\}}$ that takes into account the different effects on light polarized along the plane of incidence ($F^{\parallel}$) and light polarized perpendicular to it ($F^{\perp}$).
Here $F\in\{\mathcal{T},\mathcal{R}\}$ refers to the Fresnel transmission ($\mathcal{T}$) or reflection ($\mathcal{R}$) coefficients. 
The Fresnel Mueller matrix is given by
\begin{equation}
\label{eq:fresnelmat}
\mathbf{F}^{F\in\{\mathcal{T},\mathcal{R}\}}=
\left[
\begin{array}{cccc}
\frac{F^{\perp}+F^{\parallel}}{2} & \frac{F^{\perp}-F^{\parallel}}{2} & 0 & 0 \\
\frac{F^{\perp}-F^{\parallel}}{2} & \frac{F^{\perp}+F^{\parallel}}{2} & 0 & 0 \\
0 & 0 & \sqrt{F^{\perp}F^{\parallel}}\cos\delta & \sqrt{F^{\perp}F^{\parallel}}\sin\delta \\
0 & 0 & -\sqrt{F^{\perp}F^{\parallel}}\sin\delta & \sqrt{F^{\perp}F^{\parallel}}\cos\delta
\end{array}
\right],
\end{equation}
where $\delta$ is the retardation phase shift. The value of $\delta$ is 0 when the incident angle is larger than the Brewster angle, and $\pi$ otherwise. 

The Fresnel coefficients for reflection and transmission, denoted as $\mathcal{R}^{\perp}$, $\mathcal{R}^{\parallel}$, $\mathcal{T}^{\perp}$, and $\mathcal{T}^{\parallel}$, can be calculated as 
\begin{equation}
\mathcal{R}^{\perp}=\left(\frac{\eta_{1}\cos\theta_{1}-\eta_{2}\cos\theta_{2}}{\eta_{1}\cos\theta_{1}+\eta_{2}\cos\theta_{2}}\right)^{2} ,\mathcal{R}^{\parallel}=\left(\frac{\eta_{1}\cos\theta_{2}-\eta_{2}\cos\theta_{1}}{\eta_{1}\cos\theta_{2}+\eta_{2}\cos\theta_{1}}\right)^{2},
\end{equation}
\begin{equation}
\mathcal{T}^{\perp}=\left(\frac{2\eta_{1}\cos\theta_{1}}{\eta_{1}\cos\theta_{1}+\eta_{2}\cos\theta_{2}}\right)^{2} ,\mathcal{T}^{\parallel}=\left(\frac{2\eta_{1}\cos\theta_{1}}{\eta_{1}\cos\theta_{2}+\eta_{2}\cos\theta_{1}}\right)^{2}.
\end{equation}
These coefficients describe the polarization state of light after being reflected or transmitted at an interface, and depend on the refractive indices of the media on either side of the interface ($\eta_{1}$ and $\eta_{2}$) as well as the incident ($\theta_{1}$) and exitant ($\theta_{2}$) angles. 
We also define $\mathcal{T}^{+}=(\mathcal{T}^{\perp}+\mathcal{T}^{\parallel})/2$ and $\mathcal{T}^{-}=(\mathcal{T}^{\perp}-\mathcal{T}^{\parallel})/2$ using the Fresnel transmittance coefficients, respectively.

\newcommand{\cmk}{\color{darkgreen}{\cmark}}
\newcommand{\ycmk}{\color{cyan}{\cmark}}
\newcommand{\fmk}{---}
\begin{table*}[tbp]
	\centering
	\caption{\label{tb:relatedworks} 
		Comparison with other face capture methods. A green check mark indicates that the component is acquired by the corresponding method. 
		While many dynamic face acquisition methods obtain specular albedo by leveraging polarized light, none of them can obtain a polarimetric BSSRDF parametrization, including the linear polarization components of reflectance. Furthermore, our method is the first to measure all five listed BSSRDF parameters simultaneously with the biophysical parameters of dynamic human faces. LP stands for linear polarization filter. BP stands for bandpass filter for multispectral acquisition. Cyan check marks on the diffuse column means the method does not explicitly model the diffuse appearance. On the dynamic column, the cyan check marks mean the method does not perform tracking. In the specular and subsurface scattering columns, cyan check marks indicate the use of global (or fixed) parameters for humans. 
		}
	\vspace{-3mm}	
	\footnotesize%
	\resizebox{1.0\linewidth}{!}{
	\begin{tabular}
		{m{0.015\linewidth} | m{0.19\linewidth} | M{0.06\linewidth} M{0.04\linewidth} | M{0.07\linewidth}  M{0.05\linewidth} M{0.07\linewidth} | M{0.07\linewidth} M{0.09\linewidth} |	M{0.08\linewidth} M{0.07\linewidth}	M{0.075\linewidth} M{0.075\linewidth}M{0.075\linewidth}M{0.07\linewidth}}
		\thickhline
		\multicolumn{2}{c|}{\multirow{3}{*}{Method}}  &\multirow{3}{*}{Camera}&\multirow{3}{*}{Filter}& \multirow{3}{*}{Geometry}	& \multirow{3}{*}{Diffuse} &  \multirow{3}{*}{\shortstack{Dynamic}} & 	\multicolumn{2}{c|}{Polarization} & \multicolumn{6}{c}{Face BSSRDF parameters}\\
		\cline{8-15}
		\multicolumn{2}{c|}{}&\multicolumn{2}{c|}{}&\multicolumn{1}{c}{}&\multicolumn{1}{c}{}&\multicolumn{1}{c|}{}&Polarized light	& Polarimetric reflectance &Biophysical Params.	&Specular albedo	&Specular roughness	&Single scattering &Subsurface scattering  & Refractive index  \\
		\hline		
		\multirow{9}{*}{\parbox{1\linewidth}{\rotatebox[origin=t]{90}{Photometric stereo}}}
		&\citet{weyrich_analysis_2006}		&RGB	&\fmk	&\cmk	&\cmk	&\fmk	&\fmk	&\fmk	&\fmk	&\cmk	&\cmk	&\fmk	&\cmk	&\fmk\\
		&\citet{ma_rapid_2007}				&RGB	&LP		&\cmk	&\cmk	&\fmk	&\cmk	&\fmk	&\fmk	&\fmk	&\fmk	&\fmk	&\fmk	&\fmk\\
		&\citet{ghosh_practical_2008}		&RGB	&LP		&\cmk	&\cmk	&\fmk	&\cmk	&\fmk	&\fmk	&\cmk	&\cmk	&\cmk	&\cmk	&\fmk\\
		&\citet{ghosh_multiview_2011}		&RGB	&LP		&\cmk	&\cmk	&\fmk	&\cmk	&\fmk	&\fmk	&\cmk	&\fmk	&\fmk	&\fmk	&\fmk\\
		&\citet{fyffe_comprehensive_2011}	&RGB	&\fmk	&\cmk	&\cmk	&\cmk	&\fmk	&\fmk	&\fmk	&\cmk	&\fmk	&\fmk	&\fmk	&\fmk\\
		&\citet{fyffe_single-shot_2015}		&RGB	&LP		&\cmk	&\cmk	&\cmk	&\cmk	&\fmk	&\fmk	&\cmk	&\fmk	&\fmk	&\fmk	&\fmk\\
		&\citet{gotardo_photogeometric_2015}&RGB	&LP		&\cmk	&\cmk	&\cmk	&\cmk	&\fmk	&\fmk	&\fmk	&\fmk	&\fmk	&\fmk	&\fmk\\
		&\citet{fyffe_near-instant_2016}	&RGB	&\fmk	&\cmk	&\cmk	&\fmk	&\fmk	&\fmk	&\fmk	&\cmk	&\cmk	&\fmk	&\fmk	&\fmk\\
		&\citet{legendre_efficient_2018}	&Mono	&LP		&\cmk	&\cmk	&\fmk	&\cmk	&\fmk	&\fmk	&\cmk	&\fmk	&\fmk	&\fmk	&\fmk\\
		\hline
		\multirow{4}{*}{\parbox{1\linewidth}{\rotatebox[origin=t]{90}{Learning}}}
		&\citet{li_learning_2020}			&RGB	&LP		&\cmk	&\cmk	&\cmk	&\cmk	&\fmk	&\fmk	&\cmk	&\fmk	&\fmk	&\fmk	&\fmk\\
		&\citet{bi_deep_2021}				&RGB	&\fmk	&\cmk	&\cmk	&\cmk	&\fmk	&\fmk	&\fmk	&\fmk	&\fmk	&\fmk	&\fmk	&\fmk\\
		&\citet{liu_rapid_2022}				&RGB	&LP		&\cmk	&\cmk	&\cmk	&\cmk	&\fmk	&\fmk	&\cmk	&\fmk	&\fmk	&\fmk	&\fmk\\
		&\citet{zhang_video-driven_2022}	&RGB	&LP		&\cmk	&\cmk	&\cmk	&\cmk	&\fmk	&\fmk	&\cmk	&\fmk	&\fmk	&\fmk	&\fmk\\
		\hline
		\multirow{7}{*}{\parbox{1\linewidth}{\rotatebox[origin=t]{90}{Biophysical}}}
		&\citet{preece_spectral_2004}		&Mono	&BP		&\fmk	&\fmk	&\fmk	&\fmk	&\fmk	&\cmk	&\fmk	&\fmk	&\fmk	&\fmk	&\fmk\\
		&\citet{donner_layered_2008}		&Mono	&BP		&\fmk	&\fmk	&\fmk	&\fmk	&\fmk	&\cmk	&\fmk	&\fmk	&\fmk	&\fmk	&\fmk\\
		&\citet{jimenez_practical_2010}		&RGB	&\fmk	&\fmk	&\ycmk	&\ycmk	&\fmk	&\fmk	&\cmk	&\fmk	&\fmk	&\fmk	&\fmk	&\fmk\\
		&\citet{alotaibi_biophysical_2017}	&RGB	&\fmk	&\cmk	&\cmk	&\fmk	&\fmk	&\fmk	&\cmk	&\fmk	&\fmk	&\fmk	&\fmk	&\fmk\\
		&\citet{gitlina_practical_2020}		&RGB	&\fmk	&\fmk	&\ycmk	&\fmk	&\fmk	&\fmk	&\cmk	&\fmk	&\fmk	&\fmk	&\fmk	&\fmk\\
		&\citet{aliaga_estimation_2022}	 	&RGB	&\fmk	&\fmk	&\ycmk	&\fmk	&\fmk	&\fmk	&\cmk	&\fmk	&\fmk	&\fmk	&\fmk	&\fmk\\
		&\citet{aliaga_hyperspectral_2023}	&RGB	&\fmk	&\fmk	&\ycmk	&\fmk	&\fmk	&\fmk	&\cmk	&\fmk	&\fmk	&\fmk	&\fmk	&\fmk\\
		\hline
		\multirow{7}{*}{\parbox{1\linewidth}{\rotatebox[origin=t]{90}{Stereo matching}}}
		&\citet{bradley_high_2010}			&RGB	&\fmk	&\cmk	&\ycmk	&\cmk	&\fmk	&\fmk	&\fmk	&\fmk	&\fmk	&\fmk	&\fmk	&\fmk\\
		&\citet{beeler_high-quality_2010}	&RGB	&\fmk	&\cmk	&\ycmk	&\ycmk	&\fmk	&\fmk	&\fmk	&\fmk	&\fmk	&\fmk	&\fmk	&\fmk\\
		&\citet{beeler_high-quality_2011}	&RGB	&\fmk	&\cmk	&\ycmk	&\cmk	&\fmk	&\fmk	&\fmk	&\fmk	&\fmk	&\fmk	&\fmk	&\fmk\\
		&\citet{gotardo_practical_2018}		&RGB	&\fmk	&\cmk	&\cmk	&\cmk	&\fmk	&\fmk	&\fmk	&\cmk	&\fmk	&\fmk	&\fmk	&\fmk\\
		&\citet{riviere_single-shot_2020}	&RGB	&LP		&\cmk	&\cmk	&\cmk	&\cmk	&\fmk	&\fmk	&\cmk	&\ycmk	&\fmk	&\ycmk	&\fmk\\
		&\citet{azinovic_high-res_2022}		&RGB	&LP		&\cmk	&\cmk	&\fmk	&\cmk	&\fmk	&\fmk	&\cmk	&\fmk	&\fmk	&\fmk	&\fmk\\
		\cline{2-15}
		&\textbf{Ours}						&Polar	&BP	&\cmk	&\cmk	&\cmk	&\cmk	&\cmk	&\cmk	&\cmk	&\cmk 	&\cmk 	&\cmk 	&\cmk\\
		\thickhline
	\end{tabular}
	}%
\end{table*}

\subsection{Coordinate Conversions in Polarization}
\label{sec:coor-conversion}
Different from conventional BRDF formulation, polarimetric rendering requires a coordinate conversion matrix $\mathbf{C}(\vartheta)$ for a given angle~$\vartheta$:
 \begin{equation}
 \label{eq:ccrot}
 \mathbf{C}(\vartheta)=
 \left[
 \begin{array}{cccc}
 1 & 0 & 0 & 0 \\
 0 & \cos2\vartheta & \sin2\vartheta & 0 \\
 0 & -\sin2\vartheta & \cos2\vartheta & 0 \\
 0 & 0 & 0 & 1
 \end{array}
 \right].
 \end{equation}
The polarimetric BRDF should be defined with respect to the coordinate systems of the incident Stokes vector and exitant Stokes vector. 
A common coordinate system often used for polarimetric BRDFs consists of three orthonormal vectors~\cite{hwang_sparse_2022}: the $z$-axis follows the direction of light propagation, 
the $y$-axis ($\bar{\mathbf{y}}_{i,o}$) is aligned with the camera up vector and the $x$-axis ($\bar{\mathbf{x}}_{i,o}$) is perpendicular to both. %
The plane of incidence of the specular lobe and single-scattering lobe is defined with respect to the halfway vector~$\mathbf{h}$ while the diffuse lobe is defined by the surface normal $\mathbf{n}$.

\subsection{Polarimetric Reflectance Model}
\label{sec:pbrdf_model}
We adopt the specular and single scattering terms of the polarimetric reflectance model from the recent state-of-the-art model by \citet{hwang_sparse_2022}.

\paragraph{Specular term} 
The polarized specular reflection $\mathbf{P}_{s}$ is defined as
\begin{equation}
\label{eq:p_s}
	\mathbf{P}_{s}=\kappa_{s}\mathbf{C}_{h\rightarrow o}(-\tilde{\varphi_{o}})\mathbf{F}^{\mathcal{R}}(\theta_{d};\eta)\mathbf{C}_{i\rightarrow h}(\tilde{\varphi}_{i}),
\end{equation}
where $\theta_{d}=\cos^{-1}(\mathbf{h}\cdot\boldsymbol{\omega}_{i})$ is the zenith angle between incident light $\boldsymbol{\omega}_{i}$ and the halfway vector $\mathbf{h}$~\cite{rusinkiewicz_new_1998}, $\mathbf{F}^{\mathcal{R}}$ is the Mueller matrix form of the Fresnel reflection coefficients $\mathcal{R}$ and $\mathbf{C}_{h\rightarrow o}(-\tilde{\varphi_{o}})$ and $\mathbf{C}_{i\rightarrow h}(\tilde{\varphi}_{i})$ are the coordinate conversion matrices.
The rotation angles are given as $\tilde{\varphi}_{i,o}=\varphi_{i,o}-\pi/2$, where $\varphi_{i,o}=\tan^{-1}((\mathbf{h}\cdot\bar{\mathbf{y}}_{i,o})/(\mathbf{h}\cdot\bar{\mathbf{x}}_{i,o}))$.
The term $\kappa_{s}=\rho_s\frac{\mathcal{D}(\theta_{h};\alpha_{s})\mathcal{G}(\theta_{i},\theta_{o};\alpha_{s})}{4(\mathbf{n}\cdot\boldsymbol{\omega}_{i})(\mathbf{n}\cdot\boldsymbol{\omega}_{o})}$ is the specular reflection term,
where $\theta_{h}=\cos^{-1}(\mathbf{h}\cdot\mathbf{n})$ is the zenith angle between the normal $\mathbf{n}$ and $\mathbf{h}$, $\mathcal{D}$~represents the GGX distribution function~\cite{walter_microfacet_2007}, $\alpha_{s}$~is specular roughness term, $\mathcal{G}$ is Smith's geometric attenuation function of shadowing/masking term~\cite{heitz_understanding_2014}, and $\rho_{s}$ is the specular albedo. %

\paragraph{Single scattering term}
The practical single scattering term, on the other hand, is defined as
\begin{equation}
\label{eq:p_ss}
\mathbf{P}_{ss} \approx \kappa_{ss}\mathbf{C}_{h\rightarrow o}(-\tilde{\varphi}_{o})\mathbf{F}^{\mathcal{R}}(\theta_{d};\eta)\mathbf{C}_{i\rightarrow h}(\tilde{\varphi}_{i}),
\end{equation}
where $\kappa_{ss}	=\rho_{ss}\frac{\mathcal{D}(\theta_{h};\alpha_{ss})\mathcal{G}(\theta_{i},\theta_{o};\alpha_{ss})}{4(\mathbf{n}\cdot\boldsymbol{\omega}_{i})(\mathbf{n}\cdot\boldsymbol{\omega}_{o})}$ is the single scattering reflection term,
and $\alpha_{ss}$ and $\rho_{ss}$ represent roughness and albedo of the single scattering term, respectively.

\paragraph{Subsurface scattering term} Refer to the main paper.%

\subsection{Human Skin Rendering with Subsurface Scattering}
\label{sec:skin-rendering}
For translucent materials, exitant radiance $L_{o}(\mathbf{x}_{o},\boldsymbol{\omega}_{o})$ is computed by convolving the incident light $L_{i}(\mathbf{x}_{i},\boldsymbol{\omega}_{i})$ with a bidirectional scattering surface reflectance distribution function (BSSRDF) $\Psi$~\cite{nicodemus_geometrical_1977}:
\begin{equation}
\label{eq:bssrdf}
L_{o}(\mathbf{x}_{o},\boldsymbol{\omega}_{o})=\int_{A}\int_{2\pi}\Psi(\mathbf{x}_{i},\boldsymbol{\omega}_{i};\mathbf{x}_{o},\boldsymbol{\omega}_{o})L_{i}(\mathbf{x}_{i},\boldsymbol{\omega}_{i})(\mathbf{n}\cdot\boldsymbol{\omega}_{i})d\boldsymbol{\omega}_{i}dA(\mathbf{x}_{i}).
\end{equation}
\citet{donner_light_2005} approximate the BSSRDF of multi-layered translucent homogeneous materials using the multipole diffusion model:
\begin{equation}
\Psi(\mathbf{x}_{i},\boldsymbol{\omega}_{i};\mathbf{x}_{o},\boldsymbol{\omega}_{o})=\frac{1}{\pi}\mathcal{T}^{+}_{i}(\boldsymbol{\omega}_{i};\eta_{i})R(||\mathbf{x}_{i}-\mathbf{x}_{o}||)\mathcal{T}^{+}_{o}(\theta_{o};\eta_{o}),
\end{equation}
where $R$ is the diffuse reflectance profile and $\mathcal{T}^{+}_{i}$ and $\mathcal{T}^{+}_{o}$ are the Fresnel transmittance at the incident point $\mathbf{x}_{i}$ and the exitant point~$\mathbf{x}_{o}$. 

Given the absorption coefficients $\sigma_{a}$, reduced scattering coefficients $\sigma^{'}_{s}$, refractive index $\eta$, and the thickness of the outer layer $d$, the multipole diffusion approximation gives the forward reflectance profile $R_{\text{out}}^{{\textrm{f}}}$ and forward transmittance profile $T_{\text{out}}^{{\textrm{f}}}$ of the outer layer 
\begin{equation}
	\label{eq:reflectance}
	R_{\text{out}}^{\textrm{f}}(r)=\sum\nolimits_{k=-n}^{n}\left(P(\sigma_\textrm{tr},z_{r,k})-P(\sigma_\textrm{tr},z_{v,k})\right),
\end{equation}
\begin{equation}
	\label{eq:transmittance}
	T_{\text{out}}^{\textrm{f}}(r)=\sum\nolimits_{k=-n}^{n}\left(P(\sigma_\textrm{tr},d-z_{r,k})-P(\sigma_\textrm{tr},d-z_{v,k})\right),
\end{equation}
where $z_{r,k}$ and $z_{v,k}$ are the positions of the $k$-th positive and negative point sources, respectively.
$P(\sigma,z)=\frac{\alpha'\cdot z(1+\sigma \cdot d_{z})}{4\pi d_{z}^3}e^{-\sigma \cdot d_{z}}$ is the influence by the point source.
 $d_{z}=\sqrt{r^{2}+z^{2}}$ is the distance between the surface of the object and the point source.
  $\alpha'=\sigma_{s}'/\sigma_{t}'$ is the reduced albedo, $\sigma_\textrm{tr}=\sqrt{3\sigma_{a}\sigma_{t}'}$ is the effective transport coefficient, and $\sigma_{t}'=\sigma_{a}+\sigma_{s}'$ is the reduced extinction coefficient.

By solving the boundary conditions about the extrapolated boundaries using a multipole expansion, $(2n+1)$ multipoles are placed as
\begin{equation}
	\label{eq:pos_multipole}
\begin{aligned}
	z_{r,k}&=2k(d+z_{b}(0)+z_{b}(d))+l , \cr
	z_{v,k}&=2k(d+z_{b}(0)+z_{b}(d))-l-2z_{b}(0),
\end{aligned}
\end{equation}
where $l=1/\sigma_{t}'$ is the mean free path, $z_{b}(0)=2A(0)D$ and $z_{b}(d)=2A(d)D$ are extrapolation distances at depth $z=0$ and $z=d$, respectively.
$A(0)=\frac{1+F(0)_\textrm{dr}}{1-F(0)_\textrm{dr}}$ is the change due to internal reflection at the surface, $D=1/3\sigma_{t}'$ is the diffusion constant, and $F(0)_\textrm{dr}$ is average Fresnel reflectance~\cite{egan_determination_1973}:
\begin{equation}
	F(0)_\textrm{dr}\approx
	\left\{
	\begin{array}{cc} 
		-0.4399+\frac{0.7099}{\eta(0)}-\frac{0.3319}{\eta^{2}(0)}+\frac{0.0636}{\eta^{3}(0)}, & \eta(0)<1 \\
		-\frac{1.440}{\eta^{2}(0)}+\frac{0.710}{\eta(0)}+0.668+0.0636\eta(0), & \eta(0)>1 \\
	\end{array}
	\right.
\end{equation}
where 
$\eta(0)$ is the relative refractive index over surface $z=0$.

For the backward reflectance and transmittance profiles of the outer layer,
we can simply swap the upper and lower surfaces. Forward reflectance profile at the inner layer can be computed by assuming the thickness of the layer is infinite $d=\infty$ and dipole approximation $n=0$ using Equation~\eqref{eq:reflectance}.

\paragraph{Convolutional form of rendering equation}
Computing the analytic form of bidirectional scattering reflectance is too expensive, so 
\citet{donner_layered_2008} propose an efficient method that approximates the reflectance and transmittance profiles of multi-layered heterogeneous materials by constraining the variation of parameters to be slow relative to the mean free path, which means that properties are locally homogeneous. The efficiency of this formulation of skin rendering is especially important in our iterative optimization framework.

Given the incident flux $\Phi(\mathbf{x}_{i},\boldsymbol{\omega}_{i})$ at a surface point which can be precomputed by the incident radiance in Equation~\eqref{eq:bssrdf}, we can compute the radiant emittance profile, $M(\mathbf{x}_{o})$, by convolving the incident flux $\Phi$ with the reflectance profile $R_{\mathbf{x}_{o}}$ at exitant point $\mathbf{x}_{o}$:
\begin{equation}
	\label{eq:conv_bssrdf}
	M(\mathbf{x}_{o})=\iint\Phi(\mathbf{x}_{i},\boldsymbol{\omega}_{i})R_{\mathbf{x}_{o}}(||\mathbf{x}_{o}-\mathbf{x}_{i}||)dA=\Phi*R_{\mathbf{x}_{o}}.
\end{equation}

As opposed to the homogeneous case, the convolution of layer responses in the heterogeneous model depends on the local position on the interface between the layers.
For example, at point $\mathbf{x}_{o}$, the convolution of the heterogeneous profiles of the forward transmission of the outer layer, $T_{\text{out}}^{\textrm{f}}$, and the reflectance of the inner layer, $R_{\text{in}}^{\textrm{f}}$ results in
\begin{equation}
(T_{\text{out}}^{\textrm{f}}*R_{\text{in}}^{\textrm{f}})_{\mathbf{x}_{o}}(||\mathbf{x}_{o}-\mathbf{x}||)=\int T^{\textrm{f}}_{\text{out},\mathbf{x}_{i}}(||\mathbf{x}_{i}-\mathbf{x}||)R_{\text{in},\mathbf{x}_{o}}^{\textrm{f}}(||\mathbf{x}_{o}-\mathbf{x}_{i}||) d\mathbf{x}_{i},
\end{equation}
which depends on the convolution of the profile of the second layer $R_{\text{in},\mathbf{x}_{o}}^{\textrm{f}}$ at $\mathbf{x}_{o}$ with the transmittance responses of the first layer $T_{\text{out},\mathbf{x}_{i}}^{\textrm{f}}$ over all local positions on the interface $\mathbf{x}_{i}$. Note that $R_{\text{in},\mathbf{x}_{o}}^{\textrm{f}}$ and $T_{\text{out},\mathbf{x}_{i}}^{\textrm{f}}$ are the profiles at each $\mathbf{x}_{o}$ and $\mathbf{x}_{i}$.

Finally, the heterogeneous multi-layered forward reflectance profile $R$ can be computed by accounting for the sum of multiple inter-scattering between the two heterogeneous layers:
\begin{align}
	\label{eq:face_sss}
	R = R_\text{out}^{\textrm{f}}+\sum_{i=0}^{n} T_\text{out}^{\textrm{f}}*R_\text{in}^{\textrm{f}}*[R_\text{out}^{\textrm{b}}*R_\text{in}^{\textrm{f}}]^{i}*T_{\text{out}}^{\textrm{b}}.
\end{align}
To efficiently compute the profiles, \citet{deon_efficient_2007} use the sum of separable Gaussian functions as an accurate approximation for radially symmetric profiles by minimizing the following equation:
\begin{equation}
\min_{w_{j}}\int_{0}^{\infty}r\left(\{T,R\}_{\{\textrm{in},\textrm{out}\}}^{\{\textrm{f,b}\}}(r)-\sum_{j=1}^{m}w_{j}G(v_{j},r)\right)^{2}dr,
\end{equation}
where $v_{j}$ and $w_{j}$ are the variance and weight, respectively, of the Gaussian function $G(v,r)=\frac{1}{2\pi v}e^{-r^{2}/(2v)}$.
After optimization, we can approximate our radially symmetric profiles as the sum of separable Gaussian functions:
\begin{equation}
	\label{eq:sss_sog}
	\{T,R\}^{\{\textrm{f,b}\}}_{\{\textrm{in},\textrm{out}\}}(r)\approx\sum_{j=1}^{m}w_{j}G(v_{j},r).
\end{equation}

The convolution of separable Gaussian functions can be implemented as two 1D convolutions, which is much more efficient.
We also follow Donner et al.~\shortcite{donner_layered_2008} in representing each profile using a fixed set of Gaussians, where the variance of the Gaussian sets is a power of $4^{n}v_{0}$, where the initial variance $v_{0}$ 
is $0.01^2$ mm, from the mean free path in the outer layer. This results in the following equivalence:
\begin{equation}
	\{G(v_{0}),G(v_{0})*G(v_{0})\cdots\}=\{G(v_{0}), G(4v_{0}), G(4^{2}v_{0}),\cdots\},
\end{equation}
\noindent which allows us to compute the convolution of the next wider Gaussian function from the results of the previous narrow Gaussian function.

\section{Polarimetric Imaging}
\label{sec:polar-imaging}
\subsection{Polarimetric Image Formation Detail}
\label{sec:polar-image-formation-detail}
We now describe a new coaxial image formation designed for the polarimetric BSSRDF model. 
In our system, our light sources are equipped with a linear polarizer so that the incident light is linearly polarized, with the Stokes vector being $\mathbf{s}_{i}=[1, 1, 0, 0]^{\intercal}$. 
The Stokes vector $\mathbf{s}_{o}$ reflected from a surface point can be expressed as
\begin{equation}
	\mathbf{s}_{o}=S\mathbf{P}\mathbf{s}_{i}=S
	\left[
	\begin{array}{c}
		\kappa_{s}\mathcal{R}^{+}+\kappa_{ss}\mathcal{R}^{+}+\sum\limits_{\mathbf{x}_{i}\in\mathcal{S}}^{}\rho_{sss}(\mathcal{T}^{+}\mathcal{T}^{+}-\mathcal{T}^{-}\mathcal{T}^{+}\xi)\\
		\kappa_{s}\mathcal{R}^{+}+\kappa_{ss}\mathcal{R}^{+}-\sum\limits_{\mathbf{x}_{i}\in\mathcal{S}}^{}\rho_{sss}\mathcal{T}^{-}\mathcal{T}^{+}\xi  \\
		-\sum\limits_{\mathbf{x}_{i}\in\mathcal{S}}^{}\rho_{sss}\mathcal{T}^{-}\mathcal{T}^{+}\zeta \\
		0
	\end{array}
	\right],
\end{equation}
where 
$S={(\mathbf{n}\cdot{\boldsymbol{\omega}_{i}})}/{\Gamma^{2}}$ is the shading term with attenuation,
$\Gamma$~is the distance between the light source and the surface,
$\kappa_{s}$ is the specular reflection term of $\rho_s\frac{\mathcal{D}(\theta_{h};\alpha_{s})\mathcal{G}(\theta_{i},\theta_{o};\alpha_{s})}{4(\mathbf{n}\cdot\boldsymbol{\omega}_{i})(\mathbf{n}\cdot\boldsymbol{\omega}_{o})}$, $\kappa_{ss}$ is the single scattering reflection term of $\rho_{ss}\frac{\mathcal{D}(\theta_{h};\alpha_{ss})\mathcal{G}(\theta_{i},\theta_{o};\alpha_{ss})}{4(\mathbf{n}\cdot\boldsymbol{\omega}_{i})(\mathbf{n}\cdot\boldsymbol{\omega}_{o})}$, $\xi=\cos(2\phi)$, and $\zeta=\sin(2\phi)$.

The multi-layered subsurface scattering light interaction events lead to depolarization. As a result, the difference between Fresnel transmittances for parallel and perpendicular polarized light in both incoming and outgoing directions approaches zero: $\mathcal{T}^{-}_{o}\mathcal{T}^{-}_{i}\approx0$.
In addition, a near-coaxial setup allows for convenient simplifications in our polarimetric reflectance model~\cite{baek_simultaneous_2018,hwang_sparse_2022}. 
Geometrically, a coaxial setup results in $\boldsymbol{\omega}_{i}\approx\boldsymbol{\omega}_{o}$, $\phi_{i}\approx\pi-\phi_{o}$, $\varphi_{i}\approx 2\pi-\varphi_{o}$, $\zeta_{i}\approx-\zeta_{o}$, and $\xi_{i}\approx\xi_{o}$.
In addition, the incident angle is, by definition, below the Brewster angle, so $\cos\delta=-1$ for both the specular and the single scattering terms, (where $\delta$ is the phase shift delay, $\delta=0$~when the incident angle is larger than the Brewster angle, $\delta=\pi$ otherwise).
Last, while we define Fresnel reflection coefficients as $\mathcal{R}^{+}=(\mathcal{R}^{\perp}+\mathcal{R}^{\parallel})/2$, $\mathcal{R}^{\times}=\sqrt{\mathcal{R}^{\perp} \mathcal{R}^{\parallel}}$, and $\mathcal{R}^{-}=(\mathcal{R}^{\perp}-\mathcal{R}^{\parallel})/2$ ($\mathcal{R}^{\perp}$ and $\mathcal{R}^{\parallel}$ being the perpendicular and parallel components, respectively), in a near-coaxial setup the parallel and perpendicular Fresnel reflection coefficients become very similar, thus $\mathcal{R}^{\parallel} \approx \mathcal{R}^{\perp}$; this results in $\mathcal{R}^{-}\approx0$ and $\mathcal{R}^{+}\approx\mathcal{R}^{\times}$.

Our simplified version of the Mueller matrix thus becomes
\begin{equation}
    \label{eq:simplifiedpbssrdf}
	\begin{aligned}
		\mathbf{P}\approx\left[
		\begin{array}{cccc}
			\bar\kappa_{s,ss}\mathcal{R}^{+}+\sum\limits_{\mathbf{x}_{i}\in\mathcal{S}}^{}\rho_{sss}\mathcal{T}^{++} & -\sum\limits_{\mathbf{x}_{i}\in\mathcal{S}}^{}\rho_{sss}\mathcal{T}^{-+}\xi & \sum\limits_{\mathbf{x}_{i}\in\mathcal{S}}^{}\rho_{sss}\mathcal{T}^{-+}\zeta & 0 \\
			-\sum\limits_{\mathbf{x}_{i}\in\mathcal{S}}^{}\rho_{sss}\mathcal{T}^{-+}\xi & \bar\kappa_{s,ss}\mathcal{R}^{+} & 0 & 0\\
			-\sum\limits_{\mathbf{x}_{i}\in\mathcal{S}}^{}\rho_{sss}\mathcal{T}^{-+}\zeta & 0 & -\bar\kappa_{s,ss}\mathcal{R}^{+} & 0 \\
			0 & 0 & 0 & -\bar\kappa_{s,ss}\mathcal{R}^{+}\\ 
		\end{array}
		\right],
	\end{aligned}
\end{equation}
where 
$\bar\kappa_{s,ss}=\kappa_{s}+\kappa_{ss}$ is the sum of the specular and single scattering reflection terms, 
$\mathcal{T}^{++}=\mathcal{T}^{+}\mathcal{T}^{+}$ is the multiplication of the positive Fresnel transmission coefficients, 
and $\mathcal{T}^{-+}=\mathcal{T}^{-}\mathcal{T}^{+}$ is the multiplication of the negative/positive coefficients.

We then capture the reflected light with a polarization camera that outputs the image $\mathbf{I}$ corresponding to four linear-polarization angles as
\begin{equation}
	\begin{aligned}
		\mathbf{I}&=
		\left[
		\begin{array}{c}
			I_{0} \\
			I_{90} \\
			I_{45} \\
			I_{135}
		\end{array}
		\right]
		=
		\frac{1}{2}
		\left[
		\begin{array}{cccc}
			1 & 1 & 0 & 0 \\
			1 & -1 & 0 & 0 \\
			1 & 0 & 1 & 0 \\
			1 & 0 & -1 & 0
		\end{array}
		\right]\mathbf{s}_{o} \\
		&=
		\frac{S}{2}\left[
		\begin{array}{c}
			2\kappa_{s}\mathcal{R}^{+}+2\kappa_{ss}\mathcal{R}^{+}+\sum\limits_{\mathbf{x}_{i}\in\mathcal{S}}\rho_{sss}(\mathcal{T}^{+}\mathcal{T}^{+}-2\mathcal{T}^{-}\mathcal{T}^{+}\xi) \\
			\sum\limits_{\mathbf{x}_{i}\in\mathcal{S}}\rho_{sss}\mathcal{T}^{+}\mathcal{T}^{+} \\
			\kappa_{s}\mathcal{R}^{+}+\kappa_{ss}\mathcal{R}^{+}+\sum\limits_{\mathbf{x}_{i}\in\mathcal{S}}\rho_{sss}(\mathcal{T}^{+}\mathcal{T}^{+}-\mathcal{T}^{-}\mathcal{T}^{+}\xi-\mathcal{T}^{-}\mathcal{T}^{+}\alpha) \\
			\kappa_{s}\mathcal{R}^{+}+\kappa_{ss}\mathcal{R}^{+}+\sum\limits_{\mathbf{x}_{i}\in\mathcal{S}}\rho_{sss}(\mathcal{T}^{+}\mathcal{T}^{+}-\mathcal{T}^{-}\mathcal{T}^{+}\xi+\mathcal{T}^{-}\mathcal{T}^{+}\alpha)
		\end{array}
		\right].
	\end{aligned}
\end{equation}
From the captured images $I_{0},I_{90},I_{45},I_{135}$, we extract each component of the total reflection.
First, $I_{90}$ can be used to extract the unpolarized subsurface-scattering component. We define the unpolarized subsurface scattering observation $I_{sss}$ as
\begin{equation}
	\label{eq:I_sss}
I_{sss}=S\sum_{\mathbf{x}_{i}\in\mathcal{S}}\rho_{sss}\mathcal{T}^{+}\mathcal{T}^{+}=2I_{90},
\end{equation}
where $\mathcal{S}$ is the set of the surface points $\mathbf{x}_{i}$ of the face. %

Information about the polarized subsurface scattering term can also be obtained by subtracting $I_{135}$ from $I_{45}$, which we define as a subsurface scattering polarization observation $I_{\zeta}$ as
\begin{equation}
	\label{eq:I_alpha}
	I_{\zeta}=S\sum_{\mathbf{x}_{i}\in\mathcal{S}}\rho_{sss}\mathcal{T}^{-}\mathcal{T}^{+}\zeta=I_{135}-I_{45}.
\end{equation}

Lastly, subtracting $I_{0}$ by $I_{90}$, we can obtain a combination of specular reflection, single scattering, and oriented subsurface scattering parameters. We define this combination as the specular-dominant polarization observation $I_{s}$ as
\begin{equation}
	\label{eq:I_s}
I_{s}=S(\kappa_{s}\mathcal{R}^{+}+\kappa_{ss}\mathcal{R}^{+}-\sum_{\mathbf{x}_{i}\in\mathcal{S}}\rho_{sss}\mathcal{T}^{-}\mathcal{T}^{+}\xi)=I_{0}-I_{90}.
\end{equation}

\subsection{Spectral/Geometry Calibration of the System}
\label{sec:calib-detail}

In order to capture the spectral reflectance information of the human face, we calibrate each Dolby lens transmittance and the polarized camera response function. We use a spectrometer capture device (JETI) with a white Spectralon (99\% reflectance) to first capture the transmittance of Dolby lenses by dividing the spectral distributions of the transmitted light by the original light.
For the camera response function, we select one of the polarization cameras and capture the Spectralon images, lit by LED lights equipped with a liquid crystal tunable filter (LCTF), which transmits a selected wavelength band.
First, we estimate the spectral transmittance of the LCTF filter (similarly to the Dolby lens). Then, we capture the images every 10\,nm in range 420\,nm\,--\,670\,nm.
To calibrate across polarization cameras, we capture an image by placing a sphere-shaped Spectralon at the location where the face will be captured. Then, we normalize the captured value to the predicted camera response function with light.
For color cameras, we use a color checker to calibrate the color camera by a 3$\times$3 matrix.
To calibrate our camera's intrinsic and extrinsic parameters, we use a ChArUco checkerboard. We capture multiple images of varying checkerboard poses and minimize the reprojection errors.

\section{Computation Details}
\label{sec:details}

\subsection{Computing Normals from Heights}
\label{sec:height-to-normals}
The non-unit normal vector $\tilde{\bf{n}}$ at each pixel is computed from the displacement map $H$~\cite{riviere_single-shot_2020} as
\begin{equation}
	\begin{aligned}
	{\tilde{\bf{n}}} &= (\hat s_\textrm{u}^{}{\bf{\hat t}}_\textrm{u}^{} + {{\partial H} \over {\partial \textrm{u}}}{\bf{\hat n}}) \times (\hat s_\textrm{v}^{}{\bf{\hat t}}_\textrm{v}^{} + {{\partial H} \over {\partial \textrm{v}}}{\bf{\hat n}}) \\
	&=\left[
	\begin{matrix}
		{{\bf{\hat t}}_\textrm{u}^{}} & {{\bf{\hat t}}_\textrm{v}^{}} & {{\bf{\hat n}}}
	\end{matrix}
	\right]
	\left[
	\begin{matrix}
		{\hat{s}_{\textrm{u}}} & 0 & 0 \\
		0 & {\hat{s}_{\textrm{v}}} & 0 \\
		0 & 0 & {\hat{s}_{\textrm{u}}\hat{s}_{\textrm{v}}} \\
	\end{matrix}
	\right]
	\left[
	\begin{matrix}
		{-\frac{\delta H}{\delta \textrm{u}}} \\ {-\frac{\delta H}{\delta \textrm{v}}} \\ 1
	\end{matrix}
	\right],
	\end{aligned}
\end{equation}
where $\hat{\mathbf{t}}_{\textrm{u}}$ and $\hat{\mathbf{t}}_{\textrm{v}}$ are the finite differences in $u$ and $v$ directions of the tangent vector $\mathbf{t}$ of the initial mesh. $\hat{\mathbf{n}}$ is the unit normal vector of the mesh at the pixel, $\hat{s}_{\textrm{u}}$ and $\hat{s}_{\textrm{v}}$ are the original lengths of tangent vectors of the initial mesh. We then normalize $\tilde{\bf{n}}$ to obtain a unit vector.

\section{Optimization Details}
\label{sec:optimization-details}

\subsection{Polarimetric Inverse Rendering Details}
\label{sec:p-inverse-rendering}

For the first polarimetric inverse rendering step,
specifically, we minimize the following energy function:
\begin{equation}\label{eq:p-inverse-main-loss}
\min_{\eta, \alpha_{s}, \alpha_{ss}, \rho_{s}, \rho_{ss}, \bar{\rho}_{sss}, H}\lambda_{\psi}\mathcal{L}_{\psi}+\lambda_{sss}\mathcal{L}_{sss}+\lambda_{s}\mathcal{L}_{s}+\lambda_{\phi}\mathcal{L}_{\phi}+\mathcal{L}_{\text{reg}},
\end{equation}
where $\mathcal{L}_{\psi}$ is the refractive index loss, $\mathcal{L}_{sss}$ is the subsurface scattering loss, $\mathcal{L}_{s}$ is the specular and single scattering loss, $\mathcal{L}_{\phi}$ is the azimuthal loss, and $\mathcal{L}_{\textrm{reg}}$ is the regularizer term, $\lambda_{\psi}=0.002$, $\lambda_{sss}=1$, $\lambda_{s}=1$, $\lambda_{\phi}=1$ are the weights assigned to each loss, respectively.

\paragraph{Subsurface scattering loss}
We formulate a photometric loss of subsurface scattering $\mathcal{L}_{sss}$ by evaluating the rendered subsurface scattering image $\hat{I}^{t}_{sss}$ with the captured image $I^{t}_{sss}$ at each frame~$t$ of multiview input:
\begin{equation}
\mathcal{L}_{sss}=\sum\nolimits_{t}\textrm{V}^{t}\left(\hat{I}^{t}_{sss}-I^{t}_{sss}\right)^{2},
\end{equation}
where $\textrm{V}^{t}$ is the visibility texture map at frame $t$ for each view. The visibility map $\textrm{V}^{t}$ is $1$ at the visible pixel region and $0$ otherwise.

Solving the subsurface scattering optimization problem directly is computationally expensive and ill-posed. 
And thus, as mentioned in the main paper,
we break the optimization process into two steps to make it more manageable. 
In the first step, we make the reasonable assumption that the Fresnel transmittance of human skin does not change dramatically across the surface, as the refractive index and roughness of the skin typically change smoothly over the surface. 
Based on this assumption, we approximate the subsurface scattering reflectance as $I^{t}_{sss}=S\bar{\rho}_{sss}\mathcal{T}^{+}\mathcal{T}^{+}$, where $\bar{\rho}_{sss}$ implicitly encompasses the approximated overall observation of subsurface scattering effects at the exitant point originated from multiple incident locations. 

In the following second stage, we use the optimized surface scattering reflectance $\bar{\rho}_{sss}$ and further decompose this value using our novel inverse subsurface scattering optimization method. 
This two-step approach allows us to address the complex problem of subsurface scattering optimization in a more efficient and comprehensive manner.

\paragraph{Specular and single scattering loss}
Current methods for human face skin modeling~\cite{ma_rapid_2007,ghosh_practical_2008,riviere_single-shot_2020} and pBRDF optimization~\cite{baek_simultaneous_2018,baek_polarimetric_2021,hwang_sparse_2022} often require augmentation or clustering techniques to compensate for the limited number of specular samples per texel when determining specular and single scattering parameters. 
Thanks to our stereo imaging module, we can obtain a dense set of light-view samples for each texel by merging all video sequence frames into the reference frame as participants rotate their heads. 
This approach enables a more comprehensive analysis of skin and pBRDF properties, eliminating the need for augmentation or clustering.
We formulate the specular and single scattering loss as
\begin{equation}
	\label{eq:loss_spec}
\mathcal{L}_{s}=\sum\nolimits_{t}\textrm{V}^{t}\left(\hat{I}^{t}_{s}-I^{t}_{s}\right)^{2},
\end{equation}
where $\hat{I}^{t}_{s}$ is computed using Equation~\eqref{eq:I_s} by using $\bar{\rho}_{sss}\mathcal{T}^{-}\mathcal{T}^{+}\xi$ instead of $\sum_{\mathbf{x}_{i}\in\mathcal{S}}\rho_{sss}\mathcal{T}^{-}\mathcal{T}^{+}\xi$ and $\xi=\cos(2\phi)$.

\paragraph{Refractive index loss}
The refractive index loss is particularly relevant because it globally affects appearance at multiple levels, and our work provides a spatially-varying index of refraction from images.
We adopt the refractive-index loss from \citet{hwang_sparse_2022} that formulates the degree of polarization (DoP) of the multi-layered subsurface scattering reflections $\psi=|\mathcal{T}^{-}/\mathcal{T}^{+}|$ using unpolarized subsurface scattering image $I_{sss}$, subsurface scattering polarization image $I_{\zeta}$, and specular polarization image $I_{s}$  as
\begin{equation}
	\psi=\left|{\sqrt{(I_\zeta)^{2}+(I_\xi)^{2}}}/{I_{sss}}\right|,
\end{equation}
where $I_{\xi}=I_{s}-\kappa_{s}S\mathcal{R}^{+}-\kappa_{ss}S\mathcal{R}^{+}=-S\bar{\rho}_{sss}\mathcal{T}^{-}\mathcal{T}^{+}\xi$. 
With this observed DoP, the refractive index loss term becomes
\begin{equation}
	\label{eq:loss_refrac}
\mathcal{L}_{\psi}=\sum_{t}\text{V}^{t}\left(\hat{\psi}(\eta, \theta^{t}_{o})-\psi^{t}\right)^{2},
\end{equation}
where $\hat{\psi}$ is the predicted DoP value, which can be formulated using refractive index $\eta$ and the surface zenith angle $\theta^{t}_{o}$~\cite{atkinson_recovery_2006}. The value of $\eta$ is only optimized at the static initialization stage, but this loss term is also influenced by the local geometry defined by the displacement map $H$, which is updated at every frame.

\paragraph{Azimuthal loss}
We implement the azimuthal loss of shape from polarization as proposed by \citet{hwang_sparse_2022}:
\begin{equation}
	\label{eq:loss_azimuth}
\mathcal{L}_{\phi}=\sum_{t=1}\textrm{V}^{t}\textrm{W}^{t}_{\phi}\left((\hat{I}_{\zeta}^{t} - {I}_{\zeta}^{t})^{2}+(\hat{I}_{\xi}^{t} - {I}_{\xi}^{t})^{2}\right),
\end{equation}
where $\hat{I}_{\zeta}^{t}=S\bar{\rho}_{sss}\mathcal{T}^{-}\mathcal{T}^{+}\hat{\zeta}^{t}$ and $\hat{I}_{\xi}^{t}=-S\bar{\rho}_{sss}\mathcal{T}^{-}\mathcal{T}^{+}\hat{\xi}^{t}$ denote the diffuse polarized images obtained by optimized azimuth angles  $\hat{\zeta}^{t}=\sin(2\hat{\phi}^{t})$ and $\hat{\xi}^{t}=\cos(2\hat{\phi}^{t})$ at frame $t$, respectively. 
Note that diffuse polarization can be computed as $S\bar{\rho}_{sss}\mathcal{T}^{-}\mathcal{T}^{+}=\sqrt{({I}_{\zeta}^{t})^{2}+({I}_{\zeta}^{t})^{2}}$. 
The initial geometry extracted from multi-view stereo effectively resolves the $\pi$ ambiguity of shape from polarization~\cite{atkinson_recovery_2006,kadambi_polarized_2015}. We calculate the weight matrix $\textrm{W}^{t}_{\phi}$ by determining the normalized mean value of $I_{sss}$.

\paragraph{Regularization loss}
Our regularization loss term is designed to preserve spatial and temporal consistency and is formulated as
\begin{equation}
	\mathcal{L}_{\textrm{reg}}=\lambda_{H_{\textrm{treg}}}\mathcal{L}_{H_{\textrm{treg}}}+\lambda_{H_{\textrm{sreg}}}\mathcal{L}_{H_{\textrm{sreg}}}+\lambda_{\alpha_{s}}\mathcal{L}_{\alpha_{s}}+\lambda_{\alpha_{ss}}\mathcal{L}_{\alpha_{ss}}+\lambda_{\eta}\mathcal{L}_{\eta},
\end{equation}
where $\mathcal{L}_{H_{\textrm{treg}}}$ and $\mathcal{L}_{H_{\textrm{sreg}}}$ represent temporal and spatial regularization losses for the displacement map, $\mathcal{L}_{\alpha_{s}}$, $\mathcal{L}_{\alpha_{ss}}$, and $\mathcal{L}_{\eta}$ are correspond to spatial regularization losses for specular, single scattering roughness, and refractive index, $\lambda_{H_{\textrm{treg}}}=1$, $\lambda_{H_{\textrm{sreg}}}=1000$, $\lambda_{\alpha_{s}}=200$, $\lambda_{\alpha_{ss}}=200$, $\lambda_{\eta}=400$ are the respective weights assigned to each loss.

To preserve the geometry information of our optimized mesh relative to the initial geometry, we apply a temporal regularization loss term to our displacement map:
\begin{equation}
	\label{eq:htreg}
\mathcal{L}_{H_{\textrm{treg}}}=H^{2}.
\end{equation}
For spatial smoothness, we use the Laplacian operator on the displacement map:
\begin{equation}
	\label{eq:hsreg}
\mathcal{L}_{H_{\textrm{sreg}}}=\sum_{\mathbf{x}\in\mathcal{S}}\left(\nabla^{2}H(\mathbf{x})\right)^{2},
\end{equation}
where $\mathcal{S}$ represents the valid texture region containing the human face surface.

We assume that local spatial variations in roughness on the human face are minimal, although significant differences can be observed between distinct regions. The local variation of specularity mainly originates from variations in the specular albedo. Additionally, some specific pixels may not have a sufficient number of observations to estimate the parameters. We formulate the spatial smoothness term for the refractive index and the roughness parameter of both specular and single scattering as
\begin{equation}
	\begin{aligned}
\mathcal{L}_{\alpha_{s}}&=\sum_{\mathbf{x}\in\mathcal{S}}\left(\alpha_{s}(\mathbf{x})-\bar{\alpha}_{s}(\mathbf{x})\right)^{2},\\
\mathcal{L}_{\alpha_{ss}}&=\sum_{\mathbf{x}\in\mathcal{S}}\left(\alpha_{ss}(\mathbf{x})-\bar{\alpha}_{ss}(\mathbf{x})\right)^{2},\\
\mathcal{L}_{\eta}&=\sum_{\mathbf{x}\in\mathcal{S}}\left(\eta(\mathbf{x})-\bar{\eta}(\mathbf{x})\right)^{2},
	\end{aligned}
\end{equation}
where $\bar{\eta}(\mathbf{x})$ is the average refractive index values of the neighboring pixels of $\mathbf{x}$, $\bar{\alpha}_{s}(\mathbf{x})$ and $\bar{\alpha}_{ss}(\mathbf{x})$ are the average specular and single scattering roughness values of the neighboring pixels of $\mathbf{x}$, respectively. We employ a 5$\times$5 window to calculate the average pixel value.

\subsection{Dynamic Inverse Rendering Details}
\label{sec:dynamic-optimization}

By the given roughness parameter from the static reconstruction, we solve the following energy function to estimate the other parameters in the dynamic capture per each frame~$t$:
\begin{equation}
	\min_{\rho^{t}_{s}, \rho^{t}_{ss}, \rho^{t}_{sss}, H^{t}}\tilde{\lambda}_{sss}\mathcal{L}_{sss}+\tilde{\lambda}_{s}\mathcal{L}_{s}+\tilde{\lambda}_{\phi}\mathcal{L}_{\phi}+\tilde{\mathcal{L}}_{\text{reg}},
\end{equation}
where $\mathcal{L}_{sss}$, $\mathcal{L}_{s}$, $\mathcal{L}_{\phi}$ are inherited from the static capture loss (Equation~\eqref{eq:p-inverse-main-loss}).
$\tilde{\lambda}_{sss}=1$, $\tilde{\lambda}_{s}=1$, $\tilde{\lambda}_{\phi}=0.2$ are the weights assigned to each loss, respectively.

Here, we defined a dynamic regularization term $\tilde{\mathcal{L}}_{\text{reg}}$ as
\begin{equation}
	\tilde{\mathcal{L}}_{\text{reg}}=\tilde{\lambda}_{H^{t}_{\text{treg}}}\tilde{\mathcal{L}}_{H^{t}_{\text{treg}}}+\tilde{\lambda}_{H^{t}_{\text{sreg}}}\tilde{\mathcal{L}}_{H^{t}_{\text{sreg}}}+\tilde{\lambda}_{\rho^{t}_{s}}\tilde{\mathcal{L}}_{\rho^{t}_{s}}+\tilde{\lambda}_{\rho^{t}_{ss}}\tilde{\mathcal{L}}_{\rho^{t}_{ss}}+\tilde{\lambda}_{\bar{\rho}^{t}_{sss}}\tilde{\mathcal{L}}_{\bar{\rho}^{t}_{sss}},
\end{equation}
where $\tilde{\mathcal{L}}_{H^{t}_{\text{treg}}}$ and $\tilde{\mathcal{L}}_{H^{t}_{\text{sreg}}}$ are the dynamic temporal and spatial regularization loss for displacement map which are similar to Equations~\eqref{eq:htreg} and~\eqref{eq:hsreg}, $\tilde{\mathcal{L}}_{\rho^{t}_{s}}$, $\tilde{\mathcal{L}}_{\rho^{t}_{ss}}$,  $\tilde{\mathcal{L}}_{\bar{\rho}^{t}_{sss}}$ are the temporal regularization loss term for specular, single scattering, and subsurface scattering, and $\tilde{\lambda}_{H^{t}_{\text{treg}}}=0.001$, $\tilde{\lambda}_{H^{t}_{\text{sreg}}}=500$, $\tilde{\lambda}_{\rho^{t}_{s}}=0.01$, $\tilde{\lambda}_{\rho^{t}_{ss}}=0.01$, and $\tilde{\lambda}_{\bar{\rho}^{t}_{sss}}=0.01$ are the weights assigned to each loss, respectively.

Our temporal regularization term for specular and single scattering intensity prevents flickering artifacts in the sequence:
\begin{equation}
	\tilde{\mathcal{L}}_{\rho_{s}^{t}}=\sum_{t=1}\left(\rho_{s}^{t}-{\rho}^{0}_{s}\right)^{2},
	\tilde{\mathcal{L}}_{\rho_{ss}^{t}}=\sum_{t=1}\left(\rho_{ss}^{t}-{\rho}^{0}_{ss}\right)^{2}.
\end{equation}

In short-term dynamic sequences, changes in the color of human skin are mainly caused by variations in the hemoglobin ratio, which affects the chromaticity of the skin color. We incorporated it into our temporal subsurface scattering regularization term to minimize the difference between the albedo of the static results and that of the current frame, weighted by the intensity of the albedo:
\begin{equation}
	\tilde{\mathcal{L}}_{\rho_{sss}^{t}}=\sum_{t=1}W^{t}_{\bar{\rho}_{sss}}\left(\bar{\rho}_{sss}^{t}-\bar{\rho}^{0}_{sss}\right)^{2},
\end{equation}
where $W^{t}=|\dot{\bar{\rho}}^{0}_{sss}-\dot{\bar{\rho}}^{t}_{sss}|$ is the weight map which is computed by the difference between the intensity of the average subsurface scattering in the static results $\dot{\bar{\rho}}^{0}_{sss}$ and the intensity of the average subsurface scattering in the current frame $\dot{\bar{\rho}}^{t}_{sss}$. Finally, using the estimated average subsurface scattering reflectance $\bar{\rho}_{sss}^{t}$ at the frame, we optimize the face parameters, which are the same as the static scene reconstruction.

\subsection{Optimization of Biophysically-based Parameters Details}
\label{sec:sss-optimization}
In order to optimize biophysically-based parameters using photometric loss from rendering, we propose a coordinate descent method~\cite{wright_coordinate_2015} using alternating least squares, designed to make this optimization problem manageable. We split our optimization problem into two. The first subproblem is to obtain the spectral weights $w_j$ of the Gaussian functions from the reflectance and transmittance diffusion profiles with initial variables as
\begin{equation}
\min_{w_{j}}\int_{0}^{\infty}\left(\{T,R\}_{\{\textrm{in},\textrm{out}\}}^{\{\textrm{f,b}\}}(r)-\sum_{j=1}^{m}w_{j}G(v_{j},r)\right)^{2}dr.
\end{equation}

The second subproblem is to optimize the biophysical parameters from the spectral observation $\bar{\rho}_{sss}$ obtained from polarimetric inverse rendering with ${\rho}_{sss}$:
\begin{align}
	\label{eq:subproblem1}
	\min_{C_{\text{hd}}, C_{\text{he}}, C_{\text{m}}, \beta_{\text{m}}}\left(\bar{\rho}_{sss}-\rho_{sss}\right)^2,
\end{align}
as rendered with the approximated sum of separable Gaussians.

To render the subsurface scattering component with optimizing variables, we formulate the total reflectance $\bar{R}$ (or transmittance~$\bar{T}$) of each profile as the sum of Gaussians (SoG) as described 
in Section~\ref{sec:skin-rendering}:
\begin{align}
	\sum_{j=0}w_{i,\mathbf{x}_{o},j}&=\bar{R}_{i,\mathbf{x}_{o}}=2\pi\int_{0}^{\infty}R_{i,\mathbf{x}_{o}}(r)rdr\\ \nonumber
	&=\sum_{k=-n}^{n}\left(\text{sign}(z_{r,k})e^{-\sigma_\textrm{tr}|z_{r,k}|} - \text{sign}(z_{v,k})e^{-\sigma_\textrm{tr}|z_{v,k}|}\right), \nonumber
\end{align}
where $w_{i,\mathbf{x}_{o},j}$ is the weight of $j$-th variance of the $i$-th layer's SoG at the exitant pixel $\mathbf{x}_{o}$,
$r$ is the distance between the incident surface point and the exitant surface point,
$\text{sign}()$ is the sign function, 
and $\sigma_\textrm{tr}$ is the effective transport coefficient.
$z_{r,k}$ and $z_{v,k}$ are the positions of the $k$-th positive and negative point sources in Equation~\eqref{eq:pos_multipole}, respectively.

Using the total reflectance (or transmittance), we can rephrase the SoG by approximately convolving total reflectance using the normalized SoG as
\begin{align}
	G_{R_{i,\mathbf{x}_{o}}}(r)&=\sum_{j=0}w_{i,\mathbf{x}_{o},j}G(v_j,r)\approx\bar{R}_{i,\mathbf{x}_{o}}\sum_{j=0}\bar{w}_{i,\mathbf{x}_{o},j}G(v_j,r)\\ \nonumber
	&=\bar{R}_{i,\mathbf{x}_{o}}G_{\bar{R}_{i,\mathbf{x}_{o}}}(r),
\end{align}
where $\sum_{j=0}\bar{w}_{i,\mathbf{x}_{o},j}=1$.
We can approximate this equation similarly to texture blurring in subsurface scattering rendering methods~\cite{deon_efficient_2007,jensen_practical_2001,donner_light_2005}:
\begin{equation}
R_{i,\mathbf{x}_{o}}\approx\int\bar{R}_{i,x}G_{\bar{R}_{i,\mathbf{x}_{o}}}(||\mathbf{x}_{o}-\mathbf{x}||)d\mathbf{x}=\bar{R}_{i}*G_{\bar{R}_{i,\mathbf{x}_{o}}}.
\end{equation}

\citet{ghosh_practical_2008} measure the translucency of the human skin using a contact probe, and they show that %
it does not significantly vary spatially. 
Moreover, state-of-the-art face acquisition methods that consider the blurring due to the subsurface scattering~\cite{riviere_single-shot_2020} and the human skin rendering techniques~\cite{deon_efficient_2007,jimenez_separable_2015} also use a fixed parameter of blurriness. 

Following these observations, we assume that the level of blurriness is spatially homogeneous:
\begin{equation}
	\bar{R}_{i,\mathbf{x}_{o}}\sum_{j=0}\bar{w}_{i,\mathbf{x}_{o},j}G(v_j)\approx\bar{R}_{i,\mathbf{x}_{o}}\sum_{j=0}\bar{w}_{i,j}G(v_j)=\bar{R}_{i,\mathbf{x}_{o}}G_{\bar{R}_{i}},
\end{equation}
where $G_{\bar{R}_{i}}=\sum_{j=0}\bar{w}_{i,j}G(v_j)$.
Then, finally, we can approximate the subsurface scattering of the human skin $\tilde{R}$ as
\begin{equation}
\label{eq:simple_sss}
\tilde{R}=\bar{R}_{\textrm{out}}^{\textrm{f}}*G_{\bar{R}^{\textrm{f}}_{\textrm{out}}}+((\bar{T}_{\textrm{out}}^{\textrm{f}}*G_{\bar{T}^{\textrm{f}}_{\textrm{out}}})\cdot \bar{R}_{\textrm{in}}^{\textrm{f}}*G_{\bar{R}^{\textrm{f}}_{\textrm{in}}})\cdot \bar{T}_{\textrm{out}}^{\textrm{b}}*G_{\bar{T}^{\textrm{b}}_{\textrm{out}}}+\cdots.
\end{equation}

We use gradient descent optimization to acquire the face skin parameters.
To estimate the SoG of each profile, we first downsample the images. As the distance of the neighboring pixel becomes larger, we can ignore the spatial blurring. Then, we can render the per-pixel intensity just considering the multi-layer interaction between the total reflectance as
\begin{align}
	\bar{R}_{\mathbf{x}_{o}}=\bar{R}_{\textrm{out},\mathbf{x}_{o}}^{\textrm{f}}+\sum_{n=0} \bar{T}_{\textrm{out},\mathbf{x}_{o}}^{\textrm{f}}\bar{R}_{\textrm{in},\mathbf{x}_{o}}^{\textrm{f}}[\bar{R}_{\textrm{out},\mathbf{x}_{o}}^{\textrm{b}}\bar{R}_{\textrm{in},\mathbf{x}_{o}}^{\textrm{f}}]^{n}\bar{T}_{\textrm{out},\mathbf{x}_{o}}^{\textrm{b}}.
\end{align}

Then, we fit SoG to each diffusion profile using the median intensity pixel. 
At the original resolution, we minimize the difference between the rendered images using Equation~\eqref{eq:simple_sss} and the subsurface scattering albedo images from polarimetric inverse rendering.

\begin{table*}[tbp]
	\vspace{4mm}
	\footnotesize
	\centering
	\caption{\label{tb:res_params} 
		Estimated biophysical parameters (mean and standard deviation) of subjects on the forehead and cheek with different levels of skin tone.}
\vspace{-2mm}		
\begin{tabular}
	{M{0.012\linewidth} | m{0.1\linewidth} | M{0.14\linewidth}  M{0.14\linewidth}  M{0.14\linewidth}  M{0.14\linewidth}}
	\thickhline
	\multicolumn{6}{c}{\textbf{Forehead}} \\
	\hline
	\multicolumn{2}{c|}{Skin} & I & II & III & IV \\
	\hline
	\multicolumn{2}{c|}{Photograph}
	& \vspace{1mm} \includegraphics[width=\linewidth]{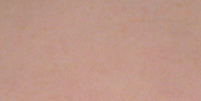}
	& \vspace{1mm} \includegraphics[width=\linewidth]{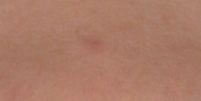}
	& \vspace{1mm} \includegraphics[width=\linewidth]{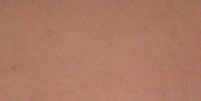}
	& \vspace{1mm} \includegraphics[width=\linewidth]{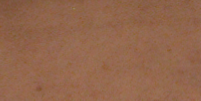} \\
	\cline{1-6}
	
	\multirow{4}{*}{\parbox{1.2\linewidth}{\rotatebox[origin=t]{90}{Skin param.}}}
	& He. (inner)		&$0.03691$ $(0.00563)$	&$0.02072$ $(0.00792)$	&$0.03622$ $(0.00903)$	&$0.02654$ $(0.00837)$ \\
	& He. (outer)		&$0.01701$ $(0.00256)$	&$0.08323$ $(0.02069)$	&$0.07337$ $(0.00441)$	&$0.10450$ $(0.01002)$ \\
	& Melanin		&$0.03243$ $(0.00183)$	&$0.05036$ $(0.00400)$	&$0.06155$ $(0.00228)$	&$0.08909$ $(0.00330)$ \\
	& Rel. eumel.		&$0.09607$ $(0.00393)$	&$0.04759$ $(0.01404)$	&$0.05304$ $(0.00658)$	&$0.05291$ $(0.00629)$ \\
	\hline
	
	\multirow{3}{*}{\parbox{1.2\linewidth}{\rotatebox[origin=t]{90}{Reflec.}}}
	& Refrac. idx		&$1.44248$ $(0.01258)$	&$1.41736$ $(0.01182)$	&$1.39902$ $(0.00759)$	&$1.41017$ $(0.00734)$ \\
	& Spec. rough.		&$0.58061$ $(0.02791)$	&$0.57004$ $(0.04008)$	&$0.54546$ $(0.02790)$	&$0.54927$ $(0.02887)$ \\
	& SS. rough.		&$0.96929$ $(0.00980)$	&$0.97965$ $(0.00731)$	&$0.95140$ $(0.01186)$	&$0.98458$ $(0.00542)$ \\
	\doublehline
	
	\multicolumn{6}{c}{\textbf{Cheek}} \\
	\hline
	\multicolumn{2}{c|}{Skin} & I & II & III & IV \\
	\hline
	\multicolumn{2}{c|}{Photograph}
	& \vspace{1mm} \includegraphics[width=\linewidth]{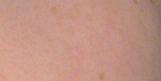}
	& \vspace{1mm} \includegraphics[width=\linewidth]{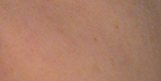}
	& \vspace{1mm} \includegraphics[width=\linewidth]{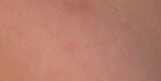}
	& \vspace{1mm} \includegraphics[width=\linewidth]{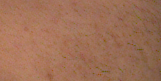}\\
	\cline{1-6}
	
	\multirow{4}{*}{\parbox{1.2\linewidth}{\rotatebox[origin=t]{90}{Skin param.}}}
	& He. (inner)		&$0.02672$ $(0.00645)$	&$0.01722$ $(0.00643)$	&$0.05753$ $(0.00970)$	&$0.02135$ $(0.01118)$ \\
	& He. (outer)		&$0.04337$ $(0.00637)$	&$0.05639$ $(0.00652)$	&$0.09753$ $(0.01057)$	&$0.09513$ $(0.01790)$ \\
	& Melanin		&$0.02891$ $(0.00286)$	&$0.04335$ $(0.00434)$	&$0.03993$ $(0.00435)$	&$0.07982$ $(0.00686)$ \\
	& Rel. eumel.		&$0.07224$ $(0.00821)$	&$0.03684$ $(0.00971)$	&$0.02355$ $(0.01014)$	&$0.05559$ $(0.01228)$ \\
	\hline
	
	\multirow{3}{*}{\parbox{1.2\linewidth}{\rotatebox[origin=t]{90}{Reflec.}}}
	& Refrac. idx		&$1.41323$ $(0.01404)$	&$1.42730$ $(0.01083)$	&$1.41453$ $(0.01454)$	&$1.42888$ $(0.01116)$ \\
	& Spec. rough.		&$0.51574$ $(0.02848)$	&$0.64520$ $(0.03452)$	&$0.53625$ $(0.02827)$	&$0.58147$ $(0.02452)$ \\
	& SS. rough.		&$0.96598$ $(0.00951)$	&$0.95688$ $(0.01927)$	&$0.97491$ $(0.00971)$	&$0.97636$ $(0.00948)$ \\
	\thickhline
\end{tabular}
\end{table*}

We use a fixed-size discrete kernel for each SoG. At each pixel $\mathbf{x}_{o}$, we first compute the distance between the neighboring pixel, which is within the kernel size, and $\mathbf{x}_{o}$. Using this distance, we can compute the discrete SoG kernel, which represents the subsurface scattering reflectance. To ensure energy conservation, we normalize the kernel.

\section{Implementation Details}
\label{subsec:imple}
\paragraph{Optimization}
To compute all the losses in each iterative optimization, 
we use a PyTorch RMSprop optimizer.
We use a 2K$\times$2K resolution texture to optimize whole parameters.
For the static initialization stage, we use 200 frames that represent different views. We implement a patch-based gradient descent optimization with a 256$\times$256 size of the patch.
For the dynamic sequence, we estimate appearance parameters per frame, and we similarly use patch-based gradient descent optimization. 
Our code runs on a machine equipped with an AMD EPYC 7763 CPU of 2.45\,GHz and a single NVIDIA A100 GPU.
For the static initialization, the polarimetric inverse rendering takes 180 minutes (150 iterations) on 200 frames (views),  
and the biophysical multispectral optimization takes 50 minutes (1,000 iterations at coarse resolution and 250 full-resolution iterations). %
Dynamic inverse rendering takes 180 minutes for 50 frames (150 iterations) in addition to the additional biophysical multispectral optimization of 20 minutes per frame (100 iterations at coarse resolution and 100 full-resolution iterations for subsurface scattering). %

\section{Additional Results}%
\label{sec:additional_results}%
In this section, we provide additional results of face acquisition.
\paragraph{Biophysical Parameters} 
We analyze the estimated biophysical parameters of the forehead and cheek areas of subjects with different levels of skin tone.
Table~\ref{tb:res_params} shows the estimated parameters as well as captured photographs.
We show that, as expected, darker skin exhibits higher concentration levels of estimated melanin. 
Moreover, the estimated refractive indices of the skin fall into the range from 1.35 to 1.55, which shows a good agreement with previous biophysical studies~\cite{anderson_optics_1981,van_gemert_skin_1989}.

\section{Additional Discussion}%
\label{sec:additional_discussions}%

\paragraph{Impact of multispectral polarimetric imaging}
Our system utilizes multiple polarimetric cameras equipped with off-the-shelf multispectral filters. Polarimetric inverse rendering enables us to separate the components of the polarimetric reflectance function. In addition, this yields a refractive index per texel. Accurate values for this refractive index are crucial for the estimation of subsurface scattering, as it disambiguates its contribution to appearance. Then, thanks to the multispectral input, we obtain concentration maps for individual biophysical components. We observe that the combination of both polarimetric and multispectral input is effective in estimating the overall range of subsurface scattering with high accuracy.

\paragraph{Spatial resolution}
We utilize a polarization camera with a spatial resolution of 2448$\times$2048. However, due to four linear polarization filters and four color filters (RGBG), its effective resolution is reduced to 612$\times$512. Although we leverage the recent proposed demosaicing algorithm~\cite{morimatsu_monochrome_2020} to enhance the spatial resolution of the images in 2K, our system's overall spatial resolution is half of that provided by conventional machine vision cameras (4K). We anticipate that the spatial resolution of BSSRDFs can be significantly improved when higher-resolution polarimetric cameras become available in the future.

\paragraph{Near-coaxial setup for polarimetric imaging}
Our coaxial imaging configuration has the potential to substantially alleviate the optimization challenges associated with polarimetric inverse rendering as evidenced by \citet{baek_simultaneous_2018} and \citet{hwang_sparse_2022}. 
However, this setup requires only one directional light to be activated when the corresponding directional camera captures the subject. 
In essence, this constraint prevents multiple polarimetric cameras at different orientations from capturing the subject simultaneously, allowing for a more efficient capture process.

\paragraph{Potential applications of photoplethysmography}
Recent progress in photoplethysmography~\cite{vilesov2022blending} allows for precise heart rate measurements by integrating a traditional RGB camera with radar signals. This represents a promising future research direction alongside our biophysical component measurements. 
However, polarization cameras, which include additional polarization filters, tend to have lower light efficiency than standard RGB cameras and often suffer from a low signal-to-noise ratio. This makes it challenging to distinguish temporal changes in skin appearance caused by heart rate fluctuations over time. As advancements continue in polarization camera technology, the prospect of using these devices for photoplethysmography presents an intriguing future research opportunity.

\begin{table}[htpb]
	\begin{center}
		\caption{\label{tab:notation}%
			Symbols and notations used in the paper.}
		\vspace{-3mm}
		\resizebox{0.82\linewidth}{!}{
			\begin{tabular}{lll} \cline{1-3}
				& Symbol					& Description  \\
				\cline{1-3}
				\multirow{10}{*}{\rotatebox{90}{Vectors/Angles}}
				& $\mathbf{n},\mathbf{h}$	& Normal/halfway vector	\\
				& $\phi_{i},\phi_{o}$	& Azimuth angle between the incident/exitant light along the plane of incidence of the normal vector \\ 
				& $\varphi_{i},\varphi_{o}$	& Azimuth angle between the incident/exitant light along the plane of the incidence of the halfway vector \\ 
				& $\theta_{i},\theta_{o},\theta_{h}$	& Zenith angle between the normal and the incident/exitant/halfway vector	\\ 
				& $\theta_{d}$							& Zenith angle between the incident light and the halfway vector	\\
				& $\mathbf{x}_{i},\mathbf{x}_{o}$				& Incident/exitant point	\\
				& $\boldsymbol{\omega}_{i},\boldsymbol{\omega}_{o}$	& Incident/exitant light direction		\\
				& $\zeta{\{i, o\}}$		& $\sin(2\phi_{\{i, o\}})$	\\
				& $\xi_{\{i, o\}}$ 		& $\cos(2\phi_{\{i, o\}})$ 	\\
				& $\delta$ & Retardation (delay) phase shift ($0$~when the incident angle is larger than the Brewster angle, $\pi$ otherwise)\\
				\cline{1-3}
				\multirow{16}{*}{\rotatebox{90}{Polarimetry}}
				& $\mathbf{s}_{i},\mathbf{s}_{o}$		& Stokes vector of the incident/exitant light to an object surface	\\
				& $\mathbf{s}_{\text{in}},\mathbf{s}_{\text{out}}$		& Stokes vector before/after the transformation event	\\
				& $I_{0},I_{90},I_{45},I_{135}$	& 0/90/45/135 degree linear polarized image 	\\
				& $I_{s},I_{sss}$		& Specular/Subsurface scattering observation image 	\\
				& $I_{\zeta}$		& Subsurface scattering polarization observation image 	\\
				& $\eta$ & Index of refraction \\
				& $\mathbf{M}$		& Mueller matrix	\\
				& $\mathbf{C}$		& Coordinate conversion matrix	\\
				& $\mathbf{D}$		& Depolarized matrix			\\
				& $\mathbf{F}^{F\in\{\mathcal{T},\mathcal{R}\}}$		& Fresnel Mueller matrix for transmission ($\mathcal{T}$) / reflection ($\mathcal{R}$)	\\
				& $\mathcal{T}^{\parallel},\mathcal{R}^{\parallel}$				& Fresnel transmission/reflection coefficient along the plane of incidence	\\
				& $\mathcal{T}^{\perp},\mathcal{R}^{\perp}$				& Fresnel transmission/reflection coefficient perpendicular to the plane of incidence	\\ 
				& $\mathcal{T}^{+},\mathcal{R}^{+}$				& $(\mathcal{T}^{\perp}+\mathcal{T}^{\parallel})/2$, $(\mathcal{R}^{\perp}+\mathcal{R}^{\parallel})/2$\\
				& $\mathcal{T}^{-},\mathcal{R}^{-}$				& $(\mathcal{T}^{\perp}-\mathcal{T}^{\parallel})/2$, $(\mathcal{R}^{\perp}-\mathcal{R}^{\parallel})/2$\\
				& $\mathbf{P}_{d}$,	$\mathbf{P}_{s}$, $\mathbf{P}_{ss}$	& Mueller matrix for diffuse/specular/single scattering reflection	\\
				& $\mathbf{P}_{sss}$	& Mueller matrix for subsurface scattering	\\
				\cline{1-3}
				\multirow{7}{*}{\rotatebox{90}{BSSRDF}}
				& $\mathcal{D},\mathcal{G}$				& Normal GGX distribution/Smith's geometric attenuation function	\\
				& $\kappa_{s},\kappa_{ss}$			& Specular/single scattering term	\\
				& $\bar{\kappa}_{s,ss}$			& $\kappa_{s}+\kappa_{ss}$	\\
				& $\alpha_{s},\alpha_{ss}$			& Roughness parameter of specular/single scattering	\\
				& $\rho_{s},\rho_{ss},\rho_{d}$		& Albedo of specular/single scattering/diffuse	\\
				& $\rho_{sss}$		& Subsurface scattering reflectance function	\\
				& $\bar\rho_{sss}$		& Averaged subsurface scattering reflectance value	\\
				\cline{1-3}
				\multirow{16}{*}{\rotatebox{90}{Subsurface scattering parameter}}
				& $\sigma_{a}^{\text{oxy}},\sigma_{a}^{\text{deoxy}}$	& Spectral absorption coefficient of oxy/deoxy hemoglobin	\\
				& $\sigma_{a}^{\text{em}},\sigma_{a}^{\text{pm}}$	& Spectral absorption coefficient of eumelanin/pheomelanin	\\
				& $\sigma_{a}^{\text{b}}$	& Spectral absorption coefficient of base human skin	\\
				& $\sigma_{a}^{\text{out}},\sigma_{a}^{\text{in}}$	& Spectral absorption coefficient of outer/inner layer \\
				& $\sigma_{s}^{\text{out}'},\sigma_{s}^{\text{in}'}$	& Reduced scattering coefficient of outer/inner layer \\
				& $\alpha'$			& Reduced albedo	\\
				& $\sigma_{t}', l$		& Reduced extinction coefficient and mean free path	\\
				& $\sigma_{\text{tr}}$	& Effective transport coefficient 	\\
				& $z_{r,k}, z_{v,k}$	& Position of the positive/negative monopole	\\
				& $D$				& Diffusion constant	\\
				& $F(0)_\text{dr}$		& Average Fresnel reflectance at the surface depth 0	\\
				& $A(0)$			& $(1+F(0)_{dr})/(1-F(0)_{dr})$	\\
				& $C_\text{h,out},C_\text{h,in}$	& Fraction of hemoglobin in outer/inner layer	\\
				& $C_\text{m}$			& Fraction of melanin in the inner layer	\\
				& $\beta_\text{m}$		& Fraction of eumelanin in the inner layer melanin	\\
				& $\gamma_\text{out},\gamma_\text{in}$	& Oxy-hemoglobin fraction in outer/inner hemoglobin	\\
				\cline{1-3}
				\multirow{13}{*}{\rotatebox{90}{Subsurface Scattering}}
				& $\Phi$	& Incident flux	\\
				& $M$		& Radiant emittance profile	\\
				& $L_{i},L_{o}$		& Incident/exitant radiance	\\
				& $\Psi$	& Bidirectional scattering-surface reflectance-distribution function \\
				\vspace{0.8mm}
				& $R_{\textrm{out}}^{\textrm{f}},T_{\textrm{out}}^{\textrm{f}}$	& Forward reflectance/transmittance profile of the outer layer	\\
				\vspace{0.8mm}
				& $R_{\textrm{out}}^{\textrm{b}},T_{\textrm{out}}^{\textrm{b}}$	& Backward reflectance/transmittance profile of the outer layer	\\
				\vspace{0.8mm}
				& $R_{\textrm{in}}^{\textrm{f}}$			& Forward reflectance profile of the inner layer	\\
				& $\bar{R}_{i},\bar{T}_{i}$		& Total reflectance/transmittance profile at layer i \\
				& $G$		& Gaussian function \\
				& $v_{j}$	& Variance of the sum of the Gaussian at index $j$ \\
				& $G_{{R}_{i}},G_{{T}_{i}}$		& Sum of Gaussian of reflectance/transmittance profile at layer $i$ \\
				& $G_{{\bar{R}}_{i}},G_{{\bar{T}}_{i}}$		& Normalized sum of Gaussian of reflectance/transmittance profile at layer $i$ \\

				\cline{1-3}
			\end{tabular}
		}%
		\vspace{0mm}%
	\end{center}
\end{table}

\bibliographystyle{ACM-Reference-Format}
\bibliography{bibliography}


\end{document}